\documentclass[conference]{IEEEtran}
\IEEEoverridecommandlockouts
\usepackage{cite}
\usepackage{amsmath,amssymb,amsfonts}
\usepackage[ruled,vlined]{algorithm2e}
\usepackage{graphicx}
\usepackage{textcomp}
\def\BibTeX{{\rm B\kern-.05em{\sc i\kern-.025em b}\kern-.08em
    T\kern-.1667em\lower.7ex\hbox{E}\kern-.125emX}}
\usepackage{cite,hhline,amsmath,amssymb,psfrag,cite,amsfonts,color}
\usepackage{epsfig}
\usepackage{epstopdf}
\usepackage{acronym}
\usepackage[table]{xcolor}
\usepackage{soul}
\usepackage{bm}
\usepackage{arydshln}
\usepackage{tikz}
\usepackage[utf8]{inputenc}
\usepackage{pgfplots} 
\usepackage{pgfgantt}
\usepackage{pdflscape}
\pgfplotsset{compat=newest} 
\pgfplotsset{plot coordinates/math parser=false}

\usepackage{graphicx}
\usepackage{subcaption}

\DeclareMathOperator*{\argmax}{arg\,max}

\acrodef{AWGN}{additive white Gaussian noise}
\acrodef{CRLB}{Cram\'er-Rao Lower Bound}
\acrodef{FIM}{Fisher Information matrix}
\acrodef{CDR}{correct detection rate}
\acrodef{CIR}{channel impulse response}
\acrodef{FAR}{false alarm rate}
\acrodef{GLRT}{generalized likelihood ratio test}
\acrodef{IS}{image similarity}
\acrodef{BS}{base station}
\acrodef{MAP}{maximum a-posteriori probability}

\acrodef{ROC}{receiver operating characteristics}
\acrodef{GP}{Gaussian process}
\acrodef{LLRT}{log-likelihood ratio test}
\acrodef{LOS}{line-of-sight}
\acrodef{SNR}{signal-to-noise ratio}
\acrodef{MDP}{Markov decision process}
\acrodef{NLOS}{non line-of-sight}
\acrodef{MLE}{maximum likelihood estimator}
\acrodef{mm-wave}{millimeter-wave}
\acrodef{PFA}{probability of false alarm}
\acrodef{POMDP}{partially observable Markov decision process}
\acrodef{RL}{reinforcement learning}
\acrodef{OG}{occupancy grid}
\acrodef{RV}{random variable}
\acrodef{RCS}{radar cross section}
\acrodef{UAV}{unmanned aerial vehicle}

\acrodef{pdf}{probability density function}

\newcommand{\statespace} {\mathcal{S}}

\newcommand{\state} {s_k}
\newcommand{\action} {a_k}

\newcommand{\futurestate} {s_{k+1}}

\newcommand{\actionspace} {\mathcal{A}}
\newcommand{\rewardspace} {\mathcal{R}}

\newcommand{\actionvaluef} {Q_{\pi}}

\newcommand{\Tmission}{T_{\mathsf{M}}}

\newcommand{\Pe}{P_{\mathsf{e}}}

\newcommand{\Pap}{P_{\mathsf{01}}}
\newcommand{\Pp}{P_{\mathsf{1}}}
\newcommand{\Ppa}{P_{\mathsf{10}}} 
\newcommand{\Pa}{P_{\mathsf{0}}}

\newcommand{\Pij}{P_{mn}}
\newcommand{\Di}{\mathcal{D}_m}
\newcommand{\Hj}{\mathcal{H}_n}
\newcommand{\Hi}{\mathcal{H}_m}

\newcommand{\Hp}{\mathcal{H}_1}
\newcommand{\Ha}{\mathcal{H}_0}
\newcommand{\Dp}{\mathcal{D}_1}
\newcommand{\Da}{\mathcal{D}_0}

\newcommand{\cm}{\mathbf{c}_i}
\newcommand{\xm}{{x}_i}
\newcommand{\ym}{{y}_i}

\newcommand{\vm}{{m}_i}
\newcommand{\Ncell}{{N}_{\mathsf{cells}}}

\newcommand{\suavk}{\mathbf{s}_{\mathsf{U}, k}}
\newcommand{\puavk}{\mathbf{p}_{\mathsf{U}, k}}

\newcommand{\puavf}{\mathbf{p}_{\mathsf{U}, k+1}}

\newcommand{\xu}{x_{\mathsf{U}, k}}
\newcommand{\yu}{y_{\mathsf{U}, k}}

\newcommand{\Deltapuk}{\Delta \mathbf{p}_{\mathsf{U}, k}}

\newcommand{\ak}{\mathbf{a}_k}
\newcommand{\Grid}{\mathcal{G}}

\newcommand{\thetab}{\theta_b}

\newcommand{\Nsteer}{N_{\textsf{rot}}}
\newcommand{\Nbin}{N_{\textsf{bins}}}
\newcommand{\Np}{N_{\textsf{p}}}

\newcommand{\Tf}{T_{\textsf{f}}}
\newcommand{\No}{N_0}

\newcommand{\fctarget}{f_{\mathsf{t}}}
\newcommand{\fcradar}{f_{\mathsf{r}}}

\newcommand{\Ted}{T_{\textsf{ED}}}

\newcommand{\ebs}{e_{bs}}

\newcommand{\Nd}{N_{\mathsf{d}}}

\newcommand{\staterv} {S_k}
\newcommand{\actionrv} {A_k}

\newcommand{\futurerewardrv} {R_{k+1}}
\newcommand{\futurereward} {r_{k+1}}
\newcommand{\futurestaterv} {S_{k+1}}

\newcommand{\hmapk}{\hat{\mathbf{m}}_{k}}

\newcommand{\map}{\mathbf{m}}
\newcommand{\zhistory}{\mathbf{e}_{1:k}}
\newcommand{\hmapik}{\hat{m}_{i, k}}

\newcommand{\mapi}{m_{i}}
\newcommand{\zk}{\mathbf{e}_k}
\newcommand{\zkpast}{\mathbf{e}_{k-1}}

\begin{document}
\title{Reinforcement Learning for UAV Autonomous Navigation, Mapping and Target Detection
}

\author{\IEEEauthorblockN{Anna Guerra}
\IEEEauthorblockA{\textit{DEI  - CNIT} \\
\textit{University of Bologna}\\
40136 Bologna, Italy \\
anna.guerra3@unibo.it}
\and
\IEEEauthorblockN{Francesco Guidi}
\IEEEauthorblockA{\textit{CNR-IEIIT} \\
40136 Bologna, Italy \\
francesco.guidi@ieiit.cnr.it}

\and
\IEEEauthorblockN{Davide Dardari}
\IEEEauthorblockA{\textit{DEI  - CNIT} \\
\textit{University of Bologna}\\
40136 Bologna, Italy \\
davide.dardari@unibo.it}
\and
\IEEEauthorblockN{Petar M. Djuri\'c}
\IEEEauthorblockA{\textit{ECE Department }\\
\textit{Stony Brook University}\\
11790 Stony Brook, New York \\
petar.djuric@stonybrook.edu}
}

\maketitle

\begin{abstract}
In this paper, we study a joint detection, mapping and navigation problem for a single unmanned aerial vehicle (UAV) equipped with a low complexity radar and flying in an unknown environment. The goal is to optimize its trajectory with the purpose of maximizing the mapping accuracy and, at the same time, to avoid areas where measurements might not be sufficiently informative from the perspective of a target detection. This problem is formulated as a Markov decision process (MDP) where the UAV is an agent that runs either a state estimator for target detection and for environment mapping, and a reinforcement learning (RL) algorithm to infer its own policy of navigation (i.e., the control law). Numerical results show the feasibility of the proposed idea, highlighting the UAV's capability of autonomously exploring areas with high probability of target detection while reconstructing the surrounding environment.
\end{abstract}


\bstctlcite{IEEEexample:BSTcontrol}

\section{Introduction}
In recent years, \acp{UAV} have gained more and more autonomy, including the capability of taking decisions based on predicting future possible scenarios and learning from past experiences.   For this reason, they are often at the center of artificial intelligence applications for both civilian and military use. They have been adopted to help rescuers during emergency situations (e.g., for care delivery or by providing a privileged point of view in search-and-rescue operations) \cite{marconi2012sherpa}, for the prevention of natural disasters \cite{merwaday2016improved}, or for increasing the safety of citizens in smart cities \cite{wang2019autonomous}.

In this context, a new challenge is to enrich such small flying robots with radar capabilities for exploring unknown environments and for tracking targets with a higher accuracy. An additional requirement is to accomplish all this by reducing the time needed to complete the mission \cite{hugler2018radar}. 
Nowadays, such operations are mainly performed by fusing all the information coming from on-board sensors as, for example, vision-based (camera or video) or inertial sensors \cite{guvenc2018detection}.
Nevertheless, such technologies fail in scarce visibility conditions or in harsh propagation environments, like indoors. 
A possible solution to overcome such shortcomings is to embed \acp{UAV} with \ac{mm-wave} radars, which due to the reduced wavelength, can be significantly miniaturized and integrated on board, and can improve the detection and tracking capabilities of targets \cite{hugler201877}. 

An autonomous \ac{UAV} with radar capabilities can be used for different applications. In particular, the scenario investigated in this paper can fit two different situations: an emergency situation where a single \ac{UAV}-radar reveals the presence of a target (e.g., a survivor who needs assistance or care delivery) in an unknown environment, and a safety situation where the same single \ac{UAV}-radar is able to detect a malicious (non-cooperative) user, hidden behind obstacles \cite{BenJam:C11,mohammed2014uavs}. In both cases, the single \ac{UAV} should rapidly decide where to explore the environment, while reconstructing a map of it and guaranteeing a robust detection of  targets.
In this direction, one possibility is the realization of dynamic radar networks by means of swarms of \acp{UAV} \cite{guerra2020dynamic,guerra2019dynamic}. This solution is more suitable for outdoor environments where a swarm of multi-agents can be deployed.
An alternative is to enhance a single \ac{UAV} with \ac{RL} capabilities for indoor autonomous navigation. 

The \ac{RL} concept has been initially proposed several decades ago with the aim of learning a control policy for maximizing a numerical reward signal \cite{sutton2018reinforcement,kuss2004gaussian}. 
According to this paradigm, an agent (e.g., a \ac{UAV}) determines which actions yield to the largest expected return while accounting for decisions based not only on the immediate reward but also rewards in the future. Operating like this, the exploring agent is capable to learn a policy while accomplishing its task in a non-myopic way.

The concept of using \ac{RL} with \acp{UAV} or radars is not new.
In \cite{ciftler2017indoor,wang2018reinforcement, jiang2019end,malhotra1997learning}, \ac{RL}-based approaches for multi-target detection problems have been proposed through the optimization of MIMO radar waveforms.  In \cite{selvi2018use}, the \ac{MDP} has been exploited for dual radar-communication problems, whereas in \cite{pham2018autonomous,liu2019reinforcement} \ac{RL} has been adopted for autonomous navigation in unknown environments, and in \cite{saxena2019optimal} for  \ac{UAV} trajectory planning. \ac{UAV} trajectory optimization using deep \ac{RL} has been also investigated in \cite{bayerlein2018trajectory,theile2020uav} with the purpose of optimizing the spectral efficiency and the communication performance while using a flying UAV-base station, whereas in \cite{ragi2013uav}, for learning policy that optimizes the agent trajectory and the tracking performance.
\begin{figure*}[t]
\psfrag{a}[lc][lc][1]{$a_k$}
\psfrag{s}[lc][lc][1]{$s_k$}
\psfrag{o}[lc][lc][1]{$o_k$}
\psfrag{r}[lc][lc][1]{$r_{k+1}\left(s_k, a_k \right)$}
\psfrag{b}[lc][lc][1]{$b_{k-1}\left({s}_k \right)$}
\psfrag{pe}[lc][lc][1]{$\hat{s}_k$}
\psfrag{ENV}[lc][lc][1]{Environment}
\psfrag{A}[lc][lc][0.9]{Agent}
\psfrag{UN}[lc][lc][0.8]{Not variable}
\psfrag{C}[lc][lc][0.8]{with the actions}
\psfrag{AF}[lc][lc][0.8]{Variable}
\psfrag{SE}[lc][lc][1]{{State} Estimator}
\psfrag{PI}[lc][lc][1]{Policy Estimator}
\psfrag{WB}[lc][lc][0.75]{(e.g., Bayesian filtering)}
\psfrag{WD}[lc][lc][0.75]{(e.g., Q-learning)}
\centerline{
\includegraphics[width=0.67\linewidth,draft=false]{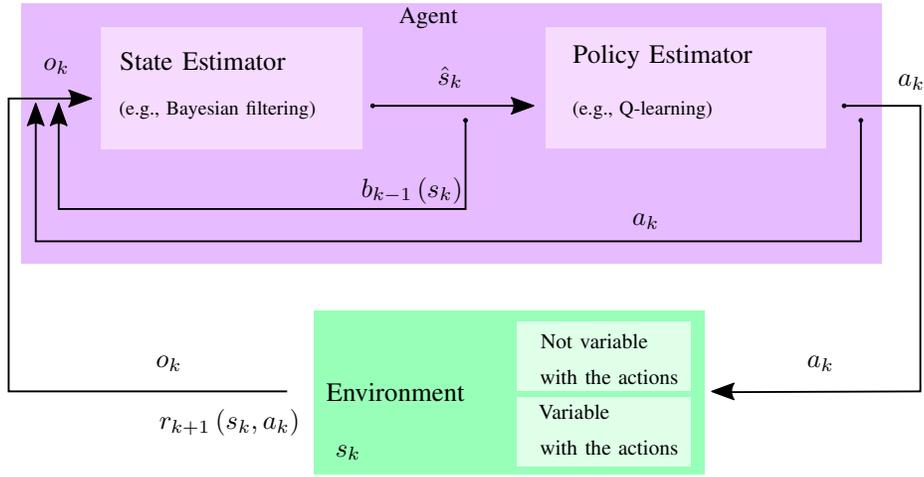}}
\caption{A belief Markov Decision Process (MDP) considered in the paper with two main blocks for state and control estimation.} 
\label{fig:POMDPscheme}
\end{figure*}

Most of these approaches, however, avoid the reconstruction of the environment (i.e., indoor mapping) whose knowledge may improve the navigation performance. 
Moreover, the mission time constraint is not always taken into account but it is essential for time-critical applications, as in emergency scenarios.

Motivated by this background, in this paper we propose a \ac{RL}-based approach for \ac{UAV} autonomous navigation and target detection. Before the navigation step, we consider a prior state estimation step that allows for inference of the map  of the environment as well as for the detection of possible targets. Then, the \ac{UAV} (the agent hereinafter)
adopts a \ac{RL} approach to learn the policy that maximizes the immediate and future rewards, expressed in the form of probability of detection error and mapping accuracy.

The rest of the paper is organized as follows. Sec.~\ref{sec:backMDP} provides a background on \ac{MDP}, whereas Sec.~\ref{sec:problem} and Sec.~\ref{sec:jointdetmap} present a description of the considered problem and the proposed solutions, respectively. Sec~\ref{sec:results} reports numerical results, and Sec.~\ref{sec:conclusions} summarizes the final conclusions.

\textbf{\textit{Notation:}} A sample space, a \ac{RV} and possible outcomes/values at time instant $k$ are indicated with $\mathcal{X}$, $X_k$, $x_k$, respectively. Vectors and matrices are denoted by bold lowercase and uppercase letters, respectively;  $p\left( x \right)$ symbolizes a probability distribution  of a random variable $x$; $p\left( x \lvert z \right)$ is the conditional distribution of $x$ given $z$; $\mathbf{x} \sim \mathcal{N}\left( \bm{\mu}, \bm{\Sigma} \right)$ means that $\mathbf{x}$ is distributed according to a  Gaussian pdf  with mean vector $\bm{\mu}$ and covariance matrix $\bm{\Sigma}$; $\mathbb{E}\left\{ \cdot \right\}$ represents the expectation of the argument; $\left[ \cdot \right]^{\mathsf{T}}$ denotes  transpose of the argument.

\section{Background on Markov Decision Processes}
\label{sec:backMDP}

The problem of learning an optimal policy to be used by an agent when exploring an environment can be formulated as a \ac{MDP}. Following the same notation as in \cite{sutton2018reinforcement}, a \ac{MDP} is defined by a tuple containing the state space  (indicated with $\statespace$), the action space (i.e., $\actionspace$), the reward space (i.e., $\rewardspace$), and the probability of transitioning from one state $\state$, at time instant $k$, to the state $\futurestate$ at time $k+1$, defined as 
\begin{align}\label{eq:transitionstates}
     p\left(\futurerewardrv=\futurereward, \futurestaterv=\futurestate \lvert \staterv=\state, \actionrv=\action \right),
\end{align}
satisfying the Markovian property.
Notably, the state at time instant $k$, indicated with $\staterv$, represents the knowledge about the environment available to the agent at time instant $k$, and can take values $\state \in \statespace$.
The actions are decided by the agent according to a specific policy given by
\begin{equation}
    \pi\left( \action \lvert \state \right) \triangleq p\left( \actionrv=\action \lvert \staterv=\state \right).
\end{equation}
The optimal policy is chosen by selecting actions that maximize a state-action value function (or Q-function), defined as 
\begin{align}\label{eq:qvalue}
&\actionvaluef\left( \state, \action \right) = \mathbb{E}_{\pi} \left\{  \sum_{\ell=0}^{\infty} \, \gamma^{\ell} \, R_{k+\ell+1}  \Big\lvert \staterv=\state, \actionrv=\action \right\}
\end{align}
where $0 \le \gamma \le 1$ is a discount rate and, for $\ell=0$, we have  the expected reward at time instant $k+1$, that is \cite{sutton2018reinforcement}
\begin{figure*}[t]
\psfrag{x}[lc][lc][0.75]{$x$}
\psfrag{y}[lc][lc][0.75]{$y$}
\psfrag{z}[lc][lc][0.75]{$z$}
\psfrag{0}[lc][lc][0.75]{$0$}
\psfrag{o1}[lc][lc][0.8]{Object 1}
\psfrag{o2}[lc][lc][0.8]{Object 2}
\psfrag{o3}[lc][lc][0.8]{Object 3}
\psfrag{t}[c][c][0.8]{\,\,\,\,\,\,\,Target}
\psfrag{a}[rc][rc][0.8]{$\mathbf{p}_0$}
\psfrag{b}[rc][rc][0.8]{UAV Trajectory}
\psfrag{v1}[lc][lc][0.8]{$\textsf{Target}$}
\psfrag{v2}[lc][lc][0.8]{$\vm=\textsf{occupied}$}
\psfrag{v3}[lc][lc][0.8]{$\vm=\textsf{empty}$}
\psfrag{o}[lc][lc][0.8]{UAV heading}
\psfrag{c}[c][c][0.75]{\qquad\qquad\qquad$\mathbf{GOAL}$: Given $\Tmission$, optimize the UAV trajectory for boosting the target detection performance}
\centerline{
\includegraphics[width=0.67\linewidth,draft=false]{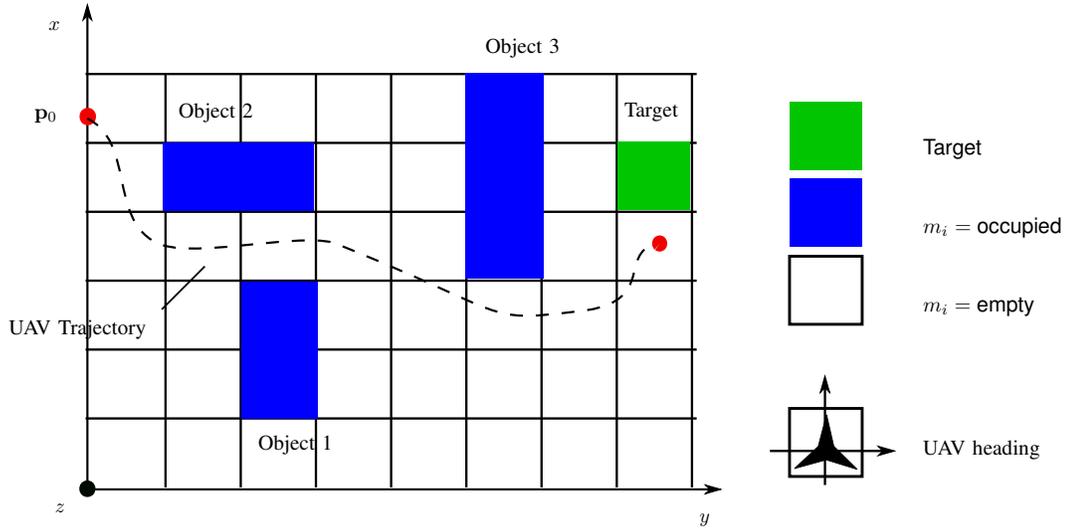} 
}
\caption{Considered UAV navigation, mapping and target detection scenario.} 
\label{fig:scenarioRL_UAV}
\end{figure*}
\begin{align}\label{eq:reward}
   & r_{k+1}(\state, \action)
    =
    \mathbb{E}\left[\futurerewardrv=\futurereward \lvert \staterv=\state, \actionrv=\action \right] \nonumber \\
    &=\sum_{\futurereward \in \rewardspace} \futurereward \sum_{\futurestate \in \statespace}  p\left(\futurereward, \futurestate \lvert \state, \action \right).
\end{align}
Consequently the problem of finding the optimal policy can be stated as
\begin{equation}\label{eq:pistar}
    \pi^*\left( \action \lvert \state \right)=\argmax_{\action} \actionvaluef\left( \state, \action \right),
\end{equation}
that can be solved iteratively using \ac{RL} approaches, e.g., temporal-difference learning \cite{sutton2018reinforcement}.

If the state is not known and can be only observed or inferred from noisy measurements, a \ac{POMDP} or Belief \ac{MDP} formalism can be used to describe the problem \cite{kaelbling1998planning,thrun2000monte}. In order to reduce the processing complexity at the agent, in the next, we will use an approach based on \ac{MDP} adopting a point estimate of the state instead of its actual value, as it will be detailed in the sequel.

In Fig.~\ref{fig:POMDPscheme}, a diagram of the system is displayed. The agent is depicted as a block where two estimation processes take place. The first is the ``\textit{State Estimator}" which can be implemented using a Bayesian filtering approach, and it provides an estimate $\hat{s}_k$ of the state ${s}_k$ starting from the previous posterior probability distribution  (indicated as $b_{k-1}\left({s}_k \right)$ in Fig.~\ref{fig:POMDPscheme}) and the current observation $o_k$. 

The second step is the ``\textit{Policy Estimator}", which infers the best action (indicated with $a_k$) to be taken to maximize the expected return. Once the agent has made a decision, it can evaluate a measure of the goodness of its behaviour by interacting with the environment in its new state. Such a measure, indicated with $r_{k+1}(s_k, a_k)$, could be expressed as a reward or a penalty, and it drives new actions of the agent. 

Some actions can change the state (e.g., by modifying the position of the agent in the space), whereas others do not alter the state of the environment (e.g., the presence of obstacles in the surrounding).

\section{Problem Statement}
\label{sec:problem}
We consider here a  target detection and mapping problem, performed by a \ac{UAV} (i.e., the agent), which autonomously navigates an indoor environment. Formally, we state the problem as follows.
\begin{itemize}
    \item[$\mathcal{P}_1$:] \textit{Detection problem}: Given a fixed maximum time to complete the mission (namely, $\Tmission$), we aim at minimizing the error in detecting a cooperative target, i.e., minimize the mis-detection and false alarm events. 
    Thus, we define
    \begin{equation}\label{eq:pe}
        \mathcal{P}_1: \min \Pe = \min \left( \Pap \, \Pp + \Ppa \,  \Pa \right),
    \end{equation}
    where $\Pij=P\left(\Di \lvert \Hj \right)$ is the probability of taking the $m$th decision $\Di$ when  the $n$th hypothesis $\Hj$ occurs, and $P_m=P\left(\Hi \right)$ is the probability of $\Hi$ being true. In our case, we have
    \begin{align}
        &\left\{\Hp, \Ha \right\}=\left\{\textsf{target}, \textsf{no target} \right\} \\
        &\left\{\Dp, \Da \right\}=\left\{\textsf{decide for $\Hp$}, \textsf{decide for $\Ha$} \right\}.
    \end{align}
 \item[$\mathcal{P}_2$:] \textit{Mapping problem}: Given a fixed maximum time to complete the mission (namely, $\Tmission$), we aim at minimizing the uncertainty in estimating the map of an unknown environment.
\end{itemize}

Hereafter, we describe the main assumptions made for solving these two joint problems. First, we consider a grid representation   $\Grid=\left\{\cm \right\}_{i=1}^{\Ncell}$ of the environment where each cell of the grid is described by a vector:
    \begin{equation}
        \cm= [\xm, \ym, \vm], \quad  i=1, \ldots, \Ncell,
    \end{equation}
    with $\Ncell$ being the number of cells with Cartesian coordinates $\left[\xm, \ym \right]$.\footnote{Note that a 2D model is used, but its 3D extension is straightforward.}
The term $\vm$ represents the \textit{state} of the cell, and we suppose that each cell can have two states, that are $\vm \in \left\{0,\, 1 \right\}=\left\{\textsf{free},\, \textsf{occupied} \right\}$. The states are unknown and should be estimated by a mapping process. For example, given the binary nature of the environment, an \ac{OG} algorithm can be used for this purpose as will be explained below.
Finally, we consider that the state of the environment does not change with time (static map and  target). 

Second, the \ac{UAV} is dynamic and, at each time instant, we define its state as
    \begin{equation}
        \suavk=\puavk=\left[ \xu, \yu\right],
    \end{equation}
    with $\puavk=\left[\xu, \yu \right] \in \Grid$ being the true Cartesian position of the \ac{UAV} at time instant $k$ constrained to lie on the 2D grid. 
    
\section{Target Detection and Mapping}
\label{sec:jointdetmap}
\begin{figure*}[t!]
\psfrag{IS}[rc][rc][0.9]{Interrogation Signal}
\psfrag{RX}[rc][rc][0.9]{Received Signal}
\psfrag{B}[rc][rc][0.9]{Backscattered Signal}
\psfrag{m}[c][c][0.9]{\quad $\hat{\mathbf{m}}_k$}
\psfrag{t}[c][c][0.9]{\quad $\hat{\mathsf{t}}_k$}
\psfrag{R}[lc][lc][0.9]{Radar}
\psfrag{RR}[lc][lc][0.9]{(TX/RX) at $\fcradar$}
\psfrag{AR}[lc][lc][0.9]{Receiver}
\psfrag{RRR}[lc][lc][0.9]{(RX) at $\fctarget$}
\psfrag{mm}[lc][lc][0.9]{\textit{Mapping module}}
\psfrag{dm}[lc][lc][0.9]{\textit{Detection module}}
\psfrag{EM}[c][c][0.9]{Energy matrix, $\mathbf{e}_k$}
\psfrag{S}[c][c][0.9]{\qquad\quad Target and Map state}
\psfrag{O}[lc][lc][0.9]{\! \!\!\!\!\! Occupancy Grid}
\psfrag{A}[lc][lc][0.9]{\!Algorithm}
\psfrag{LR}[lc][lc][0.9]{\!\!GLRT}
\psfrag{D}[lc][lc][0.9]{\!\!Energy Detector}
\psfrag{RD}[c][c][0.9]{Signal samples, $\mathbf{y}_k$}
\psfrag{SS}[c][c][0.9]{}
\centerline{
\includegraphics[width=0.75\linewidth,draft=false]{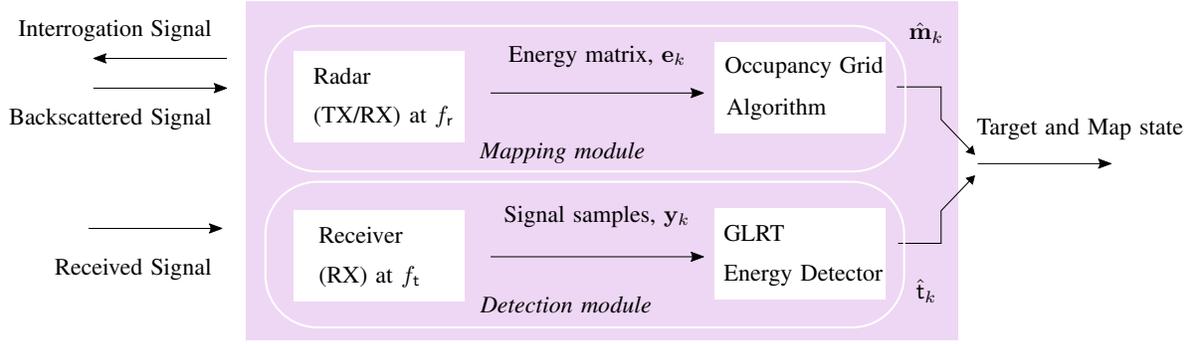} 
}
\caption{\textit{State Estimator} at the agent.} 
\label{fig:Stimatore}
\end{figure*}
In this section, we describe the main elements characterizing our  \ac{MDP} problem. The objective is to optimize the \ac{UAV} trajectory considering rewards related to the mapping and detection tasks.

\subsection{State Estimator}

The state $\mathbf{s}_k$ is a representation of the system and it consists of three subsets of states: 
\begin{enumerate}
\item The UAV position, that can be varied by the actions; 
\item  
A binary parameter indicating the presence or absence of a signal source (target) in the environment, which  cannot be changed by the actions;
\item The states of each cell, i.e., $\vm$, (not affected by  the actions). 
\end{enumerate}
Consequently, we define the state as
\begin{align}\label{eq:state}
   \mathbf{s}_k&=\left[\suavk, \mathsf{t}, m_1, \ldots, \vm, \ldots, m_{\Ncell} \right], \nonumber \\
   &=\left[\suavk, \mathsf{t}, \mathbf{m} \right],
\end{align}
where $\mathsf{t} \in \left\{ \Hp, \Ha \right\}$ denotes the presence or not of the target and $\mathbf{m}$ is the true map. The state estimate at time instant $k$ is $\hat{\mathbf{s}}_k=\left[\suavk, \hat{\mathsf{t}}_k, \hat{\mathbf{m}}_k \right]$ where $\hat{\mathsf{t}}_k \in \left\{ \Dp, \Da \right\}$ and $\hat{\mathbf{m}}_k$  are estimated by a detection and mapping modules, respectively, as represented in Fig.~\ref{fig:Stimatore}. 

The \ac{UAV} is equipped with a receiver able to process the signal coming from an active target transmitting at frequency $\fctarget$, and with a radar capable of interrogating the environment and operating at a frequency  $\fcradar$. We also assume that the overall received signals, coming from the target and backscattered by the environment, can be discerned in the frequency domain.

At each time instant $k$, the \ac{UAV} performs a scan of the environment by rotating its radar towards different directions in space. In this sense, the possibility of using large antenna arrays at \acp{mm-wave} allows for a reliable localization \cite{GueGuiDar:J18,ShaEtAl:J18,vukmirovic2019direct} and mapping performance \cite{GuiGueDar:J16} through antennas with limited size \cite{JouEtAl:J17,ghosh2019inclusive}, and, thus, being apt to be integrated even on small \acp{UAV}. To obtain reliable time-of-arrival estimates, one can choose a pulse-based radar. For each steering direction $\theta_b$, where $b=1,\,\ldots,\, \Nsteer$ is the steering index and $\Nsteer$ is the number of steering directions, the radar emits $\Np$ pulses and collects the related backscattered environment responses. At the same time, the receiver of a detection module receives the signal from the cooperative target, if the target is within the operating range of the RX.

Next, we describe the two estimation processes.

\paragraph{Occupancy Grid Mapping}
The mapping is performed using energy measurements collected by the radar from each steering direction \cite{guerra2018occupancy,GuiGueDar:J16}.
The signal received by the radar can be expressed as
\begin{align}
	r(t, \thetab)= \! \sum_{n=0}^{\Np-1} x(t-n \Tf, \thetab)+n(t),
\end{align}
where $x(t, \thetab)$ is the useful signal acquired when pointing at direction $\thetab$ and  $n(t)$ is the \ac{AWGN} with two-sided power spectral density $N_0 / 2$. 

Successively, the received signal is passed through a band-pass filter with center frequency $\fcradar$ to eliminate the out-of-band noise, thus producing a filtered signal $y(t,\thetab)$.
 
Energy measurements are then computed within a time frame $\Tf$ divided into  $\Nbin= \lfloor \Tf/\Ted  \rfloor$ discrete time bins of duration $\Ted \approx 1/W$, with $W$ being the bandwidth of the transmitted signal. Consequently, for each steering direction and for each time bin, the filtered received signal is accumulated over $\Np$ transmitted pulses so that the corresponding final energy value is given by
\begin{align}\label{eq:En1}
	\ebs\!=\!& \! \!  \sum_{n=0}^{\Np-1} \int_{(s-1) \, \Ted}^{s\, \Ted}\! y^2(t+n \Tf, \thetab) \, dt  \,,
\end{align}
\noindent with $s=1,2, \dots, \Nbin$ being the temporal bin index. 

Starting from \eqref{eq:En1} and according to the analysis in \cite{guerra2018occupancy}, the observation vector can be written as\footnote{The Gaussian approximation is valid when the number of transmitted pulses is large \cite{urkowitz1967energy}}
\begin{align}\label{eq:OBS}
    \mathbf{e}_k  &=\left[e_{11, k}, \ldots, e_{bs, k}, \ldots, e_{\Nsteer \, \Nbin, k} \right]^{\mathsf{T}} \sim \mathcal{N}\left(  \boldsymbol{\mu}_{\mathbf{e}_k} , \boldsymbol{\Sigma}_{\mathbf{e}_k}  \right),
\end{align}
where $\boldsymbol{\mu}_{\mathbf{e}_k}$ is the mean vector and $\boldsymbol{\Sigma}_{\mathbf{e}_k}$ is the measurement covariance matrix whose generic elements are given by \cite{guerra2018occupancy}
\begin{align}\label{eq:stat_ebs}
 &\mathbb{E}\left[ e_{bs, k}\right] \! =\! \Np  \int_{(s-1)\, \Ted}^{s\, \Ted} \! \!\!\! \!\!\! \tilde{x}^2(t,\thetab)\, dt + \sigma^2\, \Np \, \Ted=  E_{bs,k}+E_n \,, \nonumber \\
& \text{var}\left(  e_{bs, k} \right)= \sigma_{bs,k}^2 = \No \,\left(2\,E_{bs,k}+ E_n\right)\,,
\end{align}
where $N=\Np \Nd$ is the number of degrees of freedom, with $\Nd=2W\Ted$, $\sigma^2=N_0\, W$ is the noise power, and  $\tilde x(t)$ is the filtered version of $x(t)$.
Note that $E_{bs,k}$ depends on the backscattered response of all the map collected when the radar points towards $\thetab$ and, thus, it gathers also the energy contributions coming from all the other spatial directions filtered by the array radiation pattern. 

In particular, the considered mapping algorithm  accounts for the observation model in \eqref{eq:OBS}, and it is based on an extension of the analysis of \cite{thrun2003learning}.
Consequently, we have
\begin{align}
    & E_{bs,k}(\mathbf{m})\!\!=\!\! 
     \sum_{l \in \mathcal{I}(s)} \int_W \frac{L_\text{0}(f)\, \rho_{l}}{\left(d_{ik}\right)^4}   G^2(\tilde{\theta}_l, f)  \, df, 
\end{align}
where $\mathcal{I}(s)$ is the set of cells located at the same discrete distance $s$ from the radar, $L_\text{0}(f)=\frac{P_t(f)\, T_{\mathsf{f}}\, c^2}{f^2\, \left(4\, \pi \right)^3}$ is the path-loss at the reference distance of 1 meter, $P_t(f)$ is the power spectral density of the transmitted signal, $c$ is the speed of light, $G\left( \theta \right)$ is the array gain,  $\tilde{\theta}_l=\theta_l-\theta_b$ is the difference between the arrival and steering angles, and $\rho_{l}$ is the \ac{RCS} of the $l$th cell.

The goal of the \ac{UAV} is to infer the map of the environment by searching the maximum of the belief given the history of measurements (\ac{MAP} estimator), i.e., \cite{Thrun:C01,robbiano2019bayesian}
\begin{equation}\label{eq:ML}
\hmapk= \argmax\limits_{\map} b_k\left(\map \right) = \argmax\limits_{\map} p\left(\map \lvert \zhistory \right)\,,
\end{equation}
where $ b_k(\map)=p(\map \lvert \zhistory)$ is the posterior of the probability distribution of the map given the set of measurements collected until the discrete time $k$, i.e., $\zhistory$. 
According to \eqref{eq:ML}, the mapping problem is described as a maximum a posteriori estimation problem in a high-dimensional space, and, thus, its direct computation is prohibitive. In order to reduce the complexity, instead of computing the joint conditional probability distribution $p\left(\map \lvert \zhistory \right)$, we operate cell-by-cell as
\begin{align}\label{eq:pform}
\hmapik &= \argmax\limits_{\mapi} b_k\left(\mapi \right)=\argmax\limits_{\mapi} p\left(\mapi \lvert \zhistory \right) \nonumber \\
&=\argmax\limits_{\mapi} \frac{p\left(\zk \lvert \mapi \right)  \,  b_{k-1}\left(\mapi \right) }{p \left(\zk \lvert \zkpast \right)}.
\end{align}
Moreover, given the binary nature of cells, we can write
\begin{align}\label{eq:belief_m1}
 b_k \left( {m}_{i}=1  \right) &=\frac{  {p \left( \zk \lvert {m}_{i}=1 \right)}  \, b_{k-1}\left({m}_{i}=1 \right) }{p \left(\zk \lvert \zkpast \right)}  \\
 b_k \left( {m}_{i}=0  \right) &=1-b_k\left( \mapi=1 \right) \nonumber \\
&=\frac{  {p \left( \zk \lvert {m}_{i}=0 \right)}  \, b_{k-1}\left({m}_{i}=0 \right) }{p \left(\zk \lvert \zkpast \right)}. \label{eq:belief_m0}
\end{align}
Taking the ratio between \eqref{eq:belief_m1}-\eqref{eq:belief_m0} and considering a log-odd notation, i.e., $\ell_k\left(m_i=1\right) \triangleq \log\left(\frac{b_k(m_i=1)}{1-b_k({m}_{i}=1)} \right)$, we can obtain 
\begin{align}
    \!\!\!&\ell_k\left(m_i=1\right)\!=\!\log\left(\frac{p \left(\zk \lvert {m}_{i}=1    \right) }{{p \left( \zk \lvert {m}_{i}=0  \right)}} \right) +\ell_{k-1}\left(m_i=1 \right). 
\end{align}
\begin{figure*}[t]
\psfrag{N16E5T1}[c][c][0.9]{$N=4 \times 4$, EIRP= 5 dBm, $\mathcal{T}_1$}
\psfrag{N14E5T2}[c][c][0.9]{$N=4 \times 4$, EIRP= 5 dBm, $\mathcal{T}_2$}
\psfrag{N16E5T3}[c][c][0.9]{$N=4 \times 4$, EIRP= 5 dBm, $\mathcal{T}_3$}
\psfrag{N100E5T1}[c][c][0.9]{$N=10 \times10$, EIRP= 5 dBm, $\mathcal{T}_1$}
\psfrag{N100E5T2}[c][c][0.9]{$N=10 \times10$, EIRP= 5 dBm, $\mathcal{T}_2$}
\psfrag{N100E5T3}[c][c][0.9]{$N=10\times10$, EIRP= 5 dBm, $\mathcal{T}_3$}
\psfrag{T1}[lc][lc][0.9]{$\mathcal{T}_1$}
\psfrag{T2}[lc][lc][0.9]{$\mathcal{T}_2$}
\psfrag{T3}[lc][lc][0.9]{$\mathcal{T}_3$}
\psfrag{x}[lc][lc][0.9]{$x$ [m]}
\psfrag{y}[lc][lc][0.9]{$y$ [m]}
\centerline{
\includegraphics[width=0.35\linewidth,draft=false]{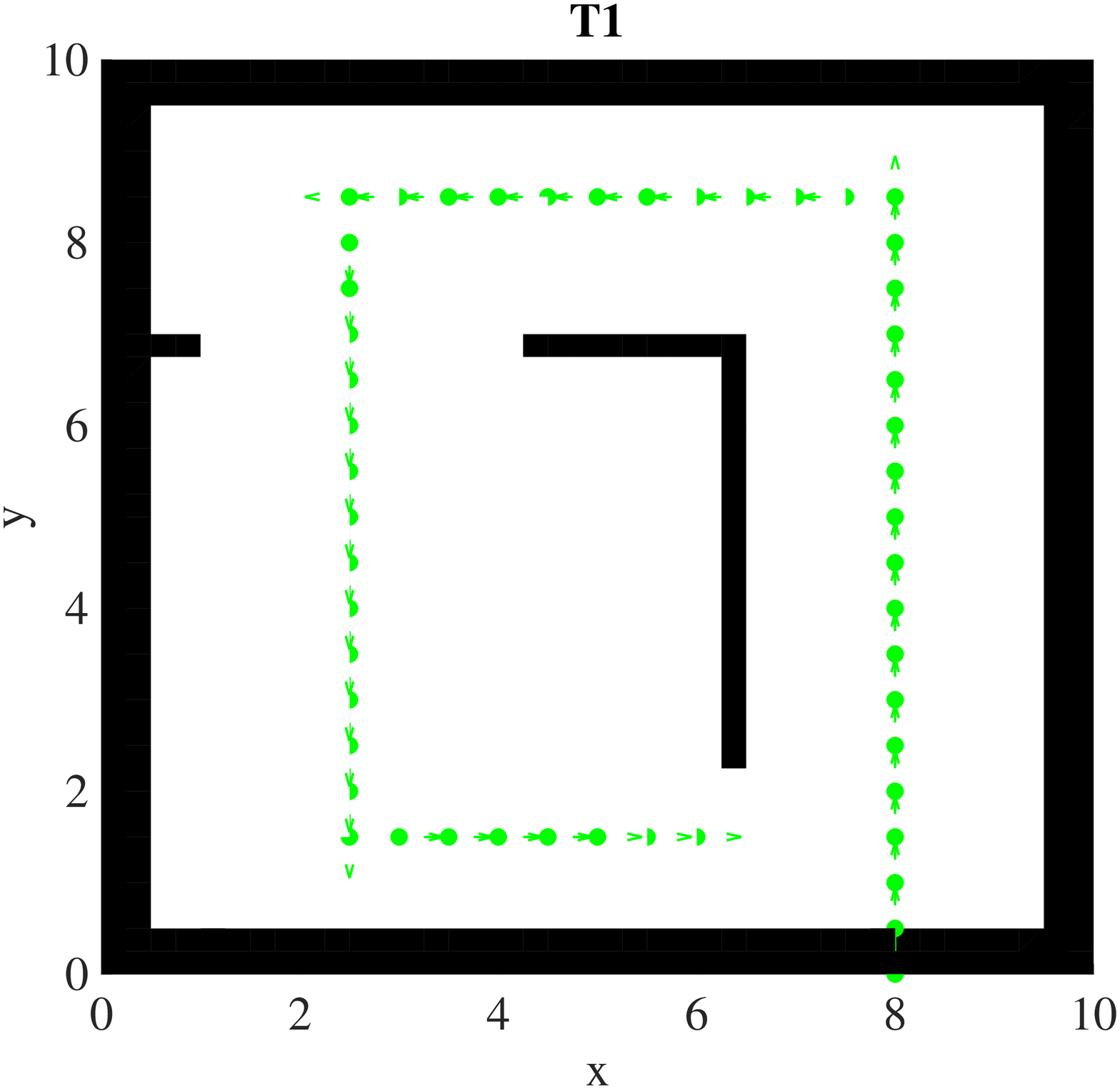} 
\includegraphics[width=0.35\linewidth,draft=false]{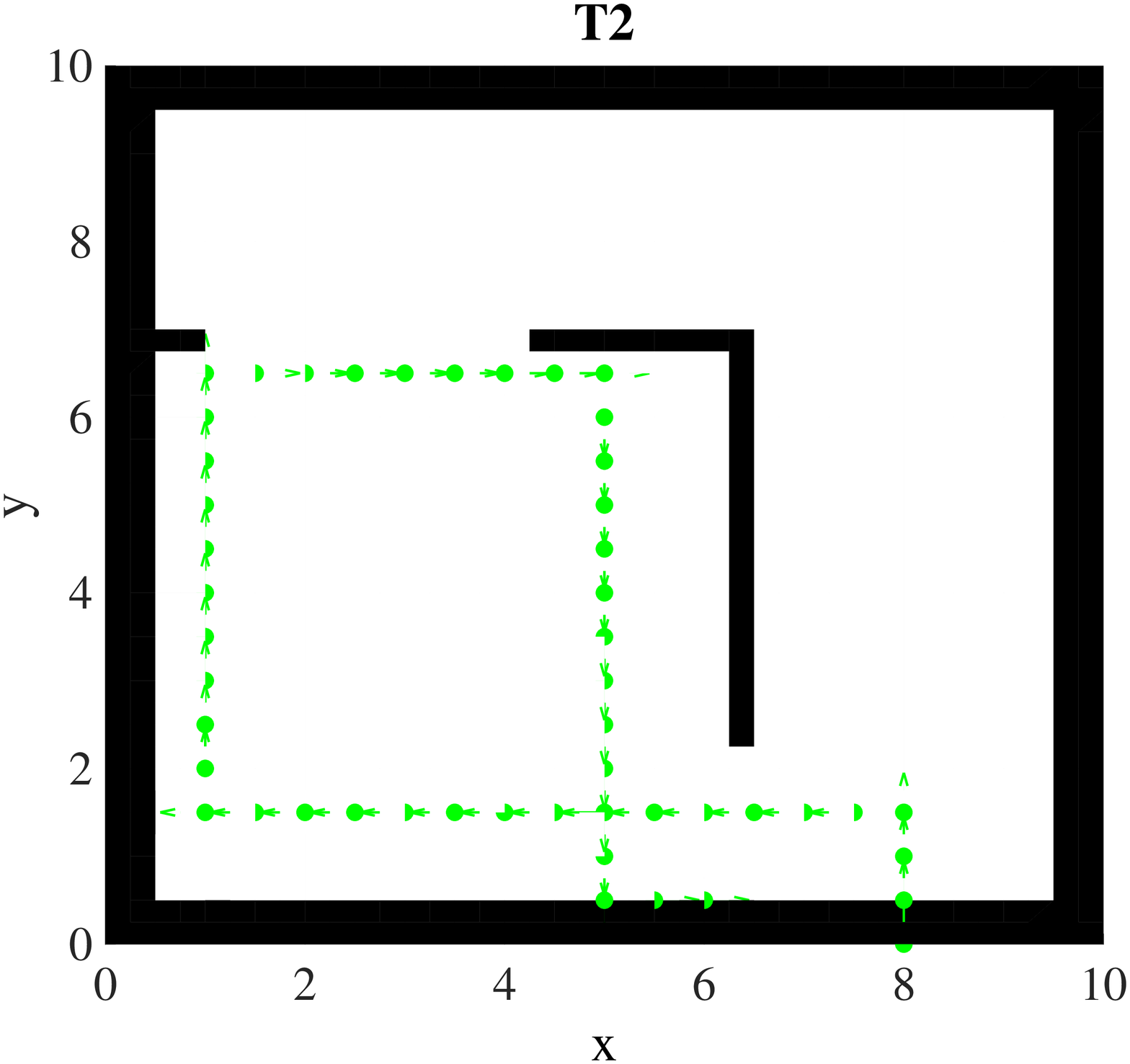} 
\includegraphics[width=0.35\linewidth,draft=false]{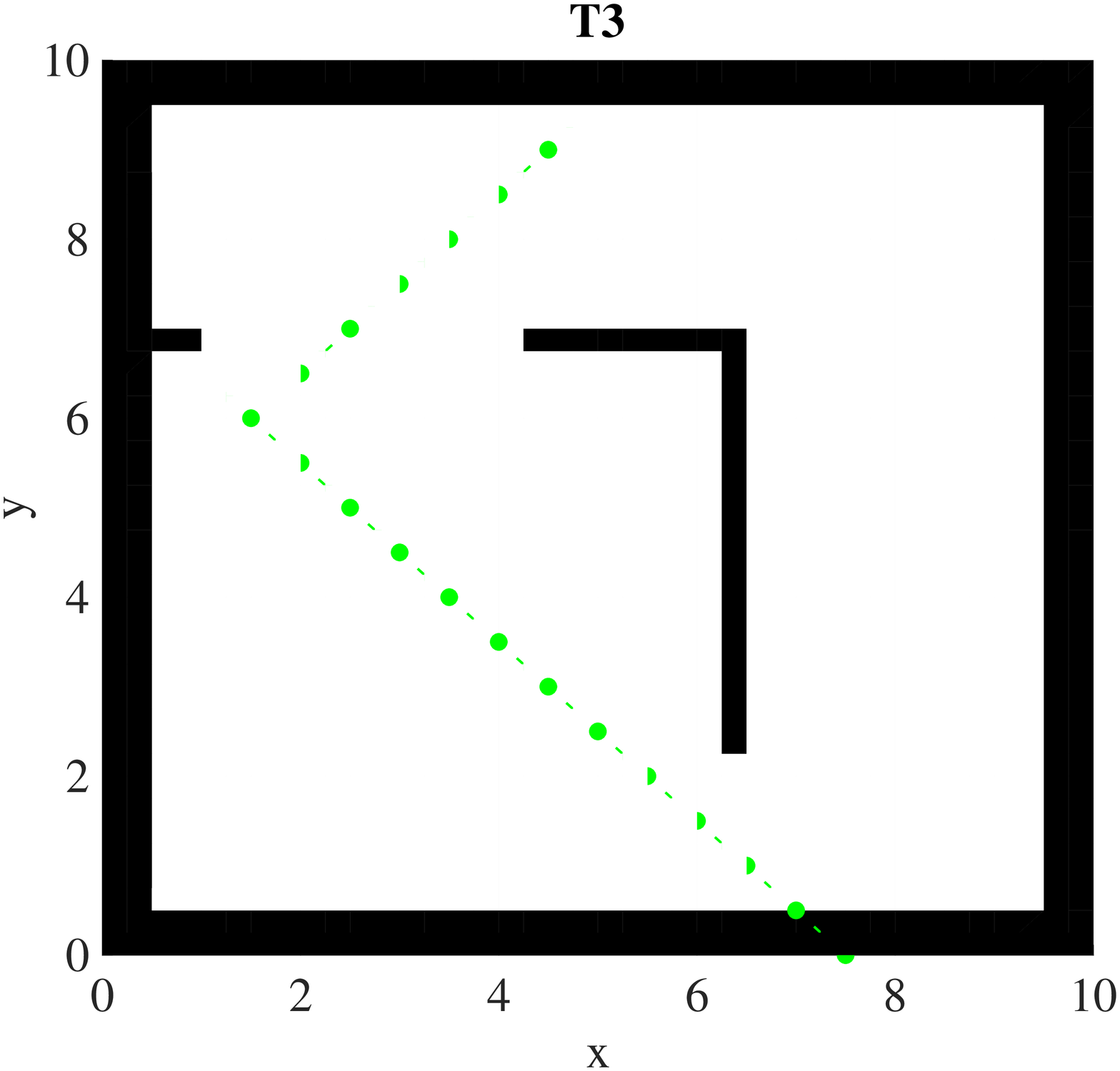}
}
\vspace{0.2cm}
\centerline{
\includegraphics[width=0.35\linewidth,draft=false]{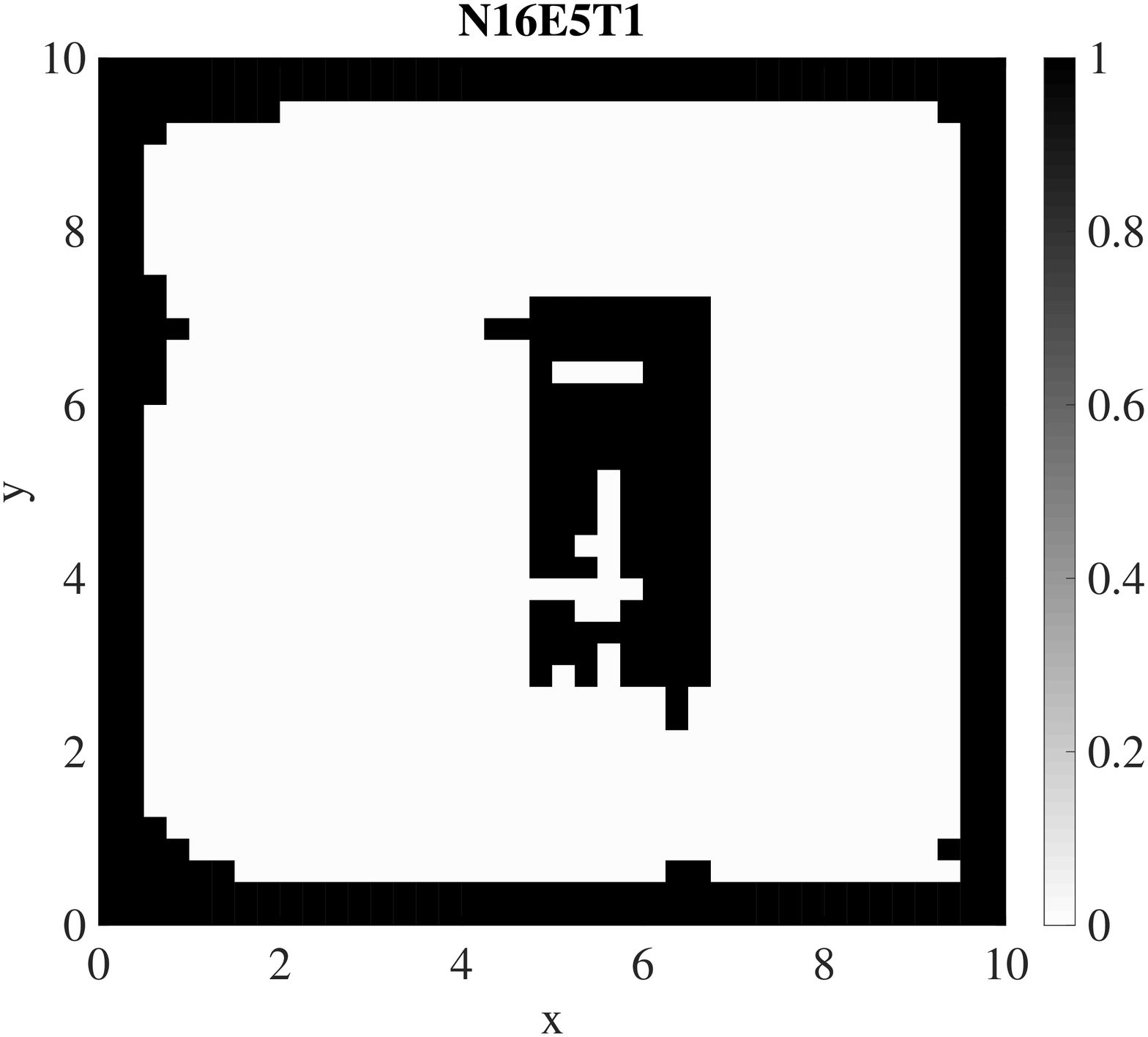} 
\includegraphics[width=0.35\linewidth,draft=false]{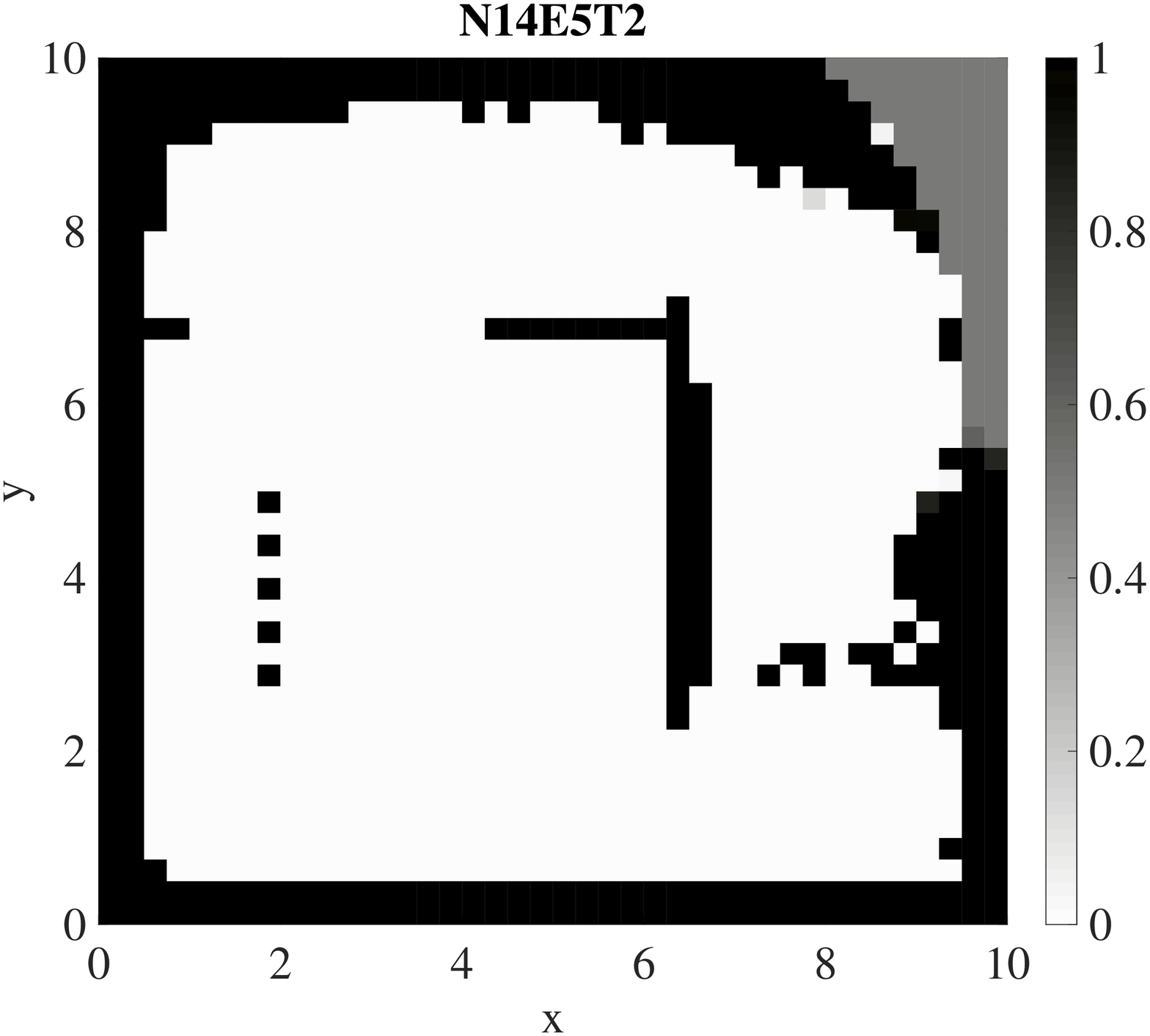} 
\includegraphics[width=0.35\linewidth,draft=false]{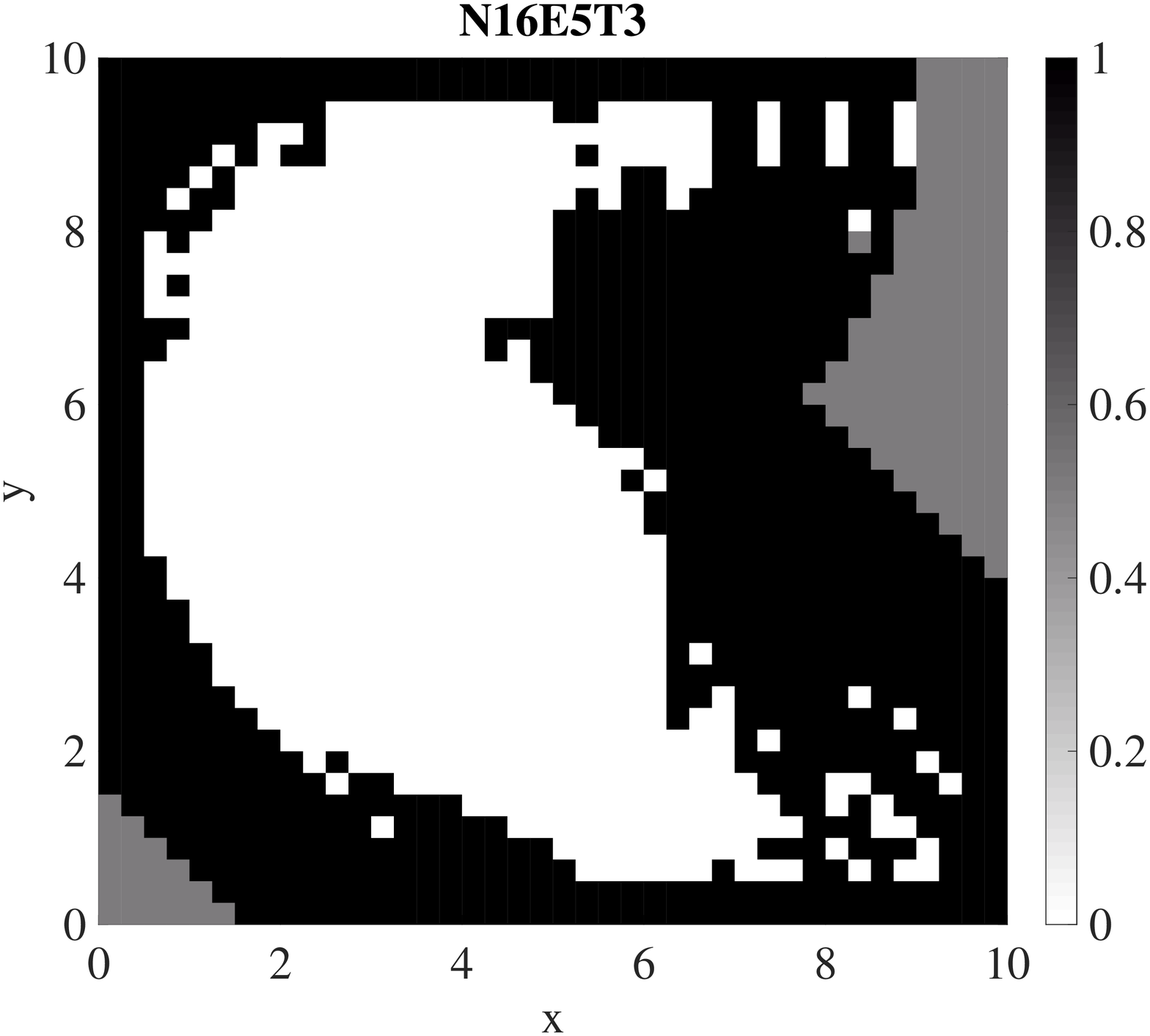}
}
\vspace{0.2cm}
\centerline{
\includegraphics[width=0.35\linewidth,draft=false]{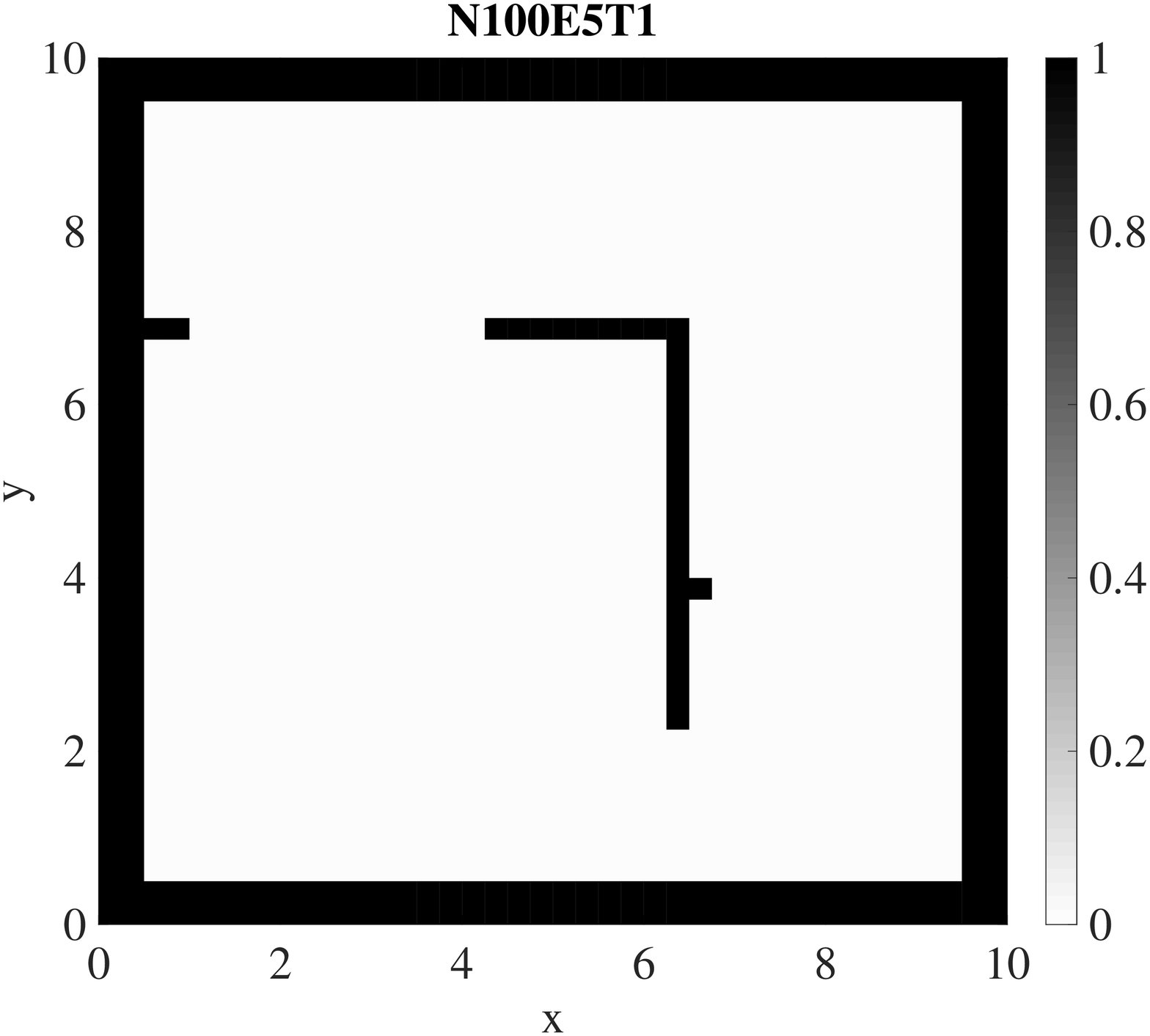} 
\includegraphics[width=0.35\linewidth,draft=false]{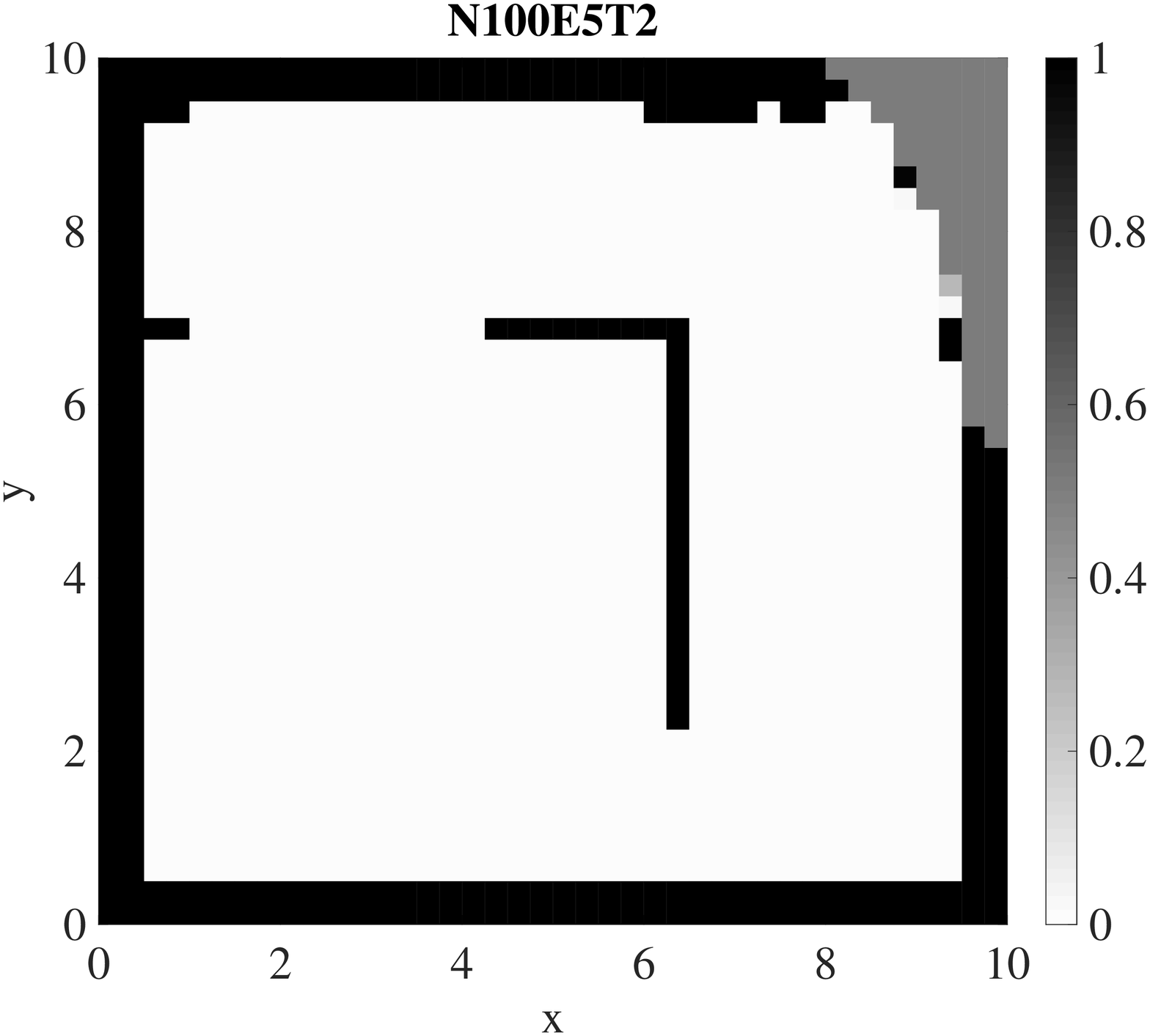} 
\includegraphics[width=0.35\linewidth,draft=false]{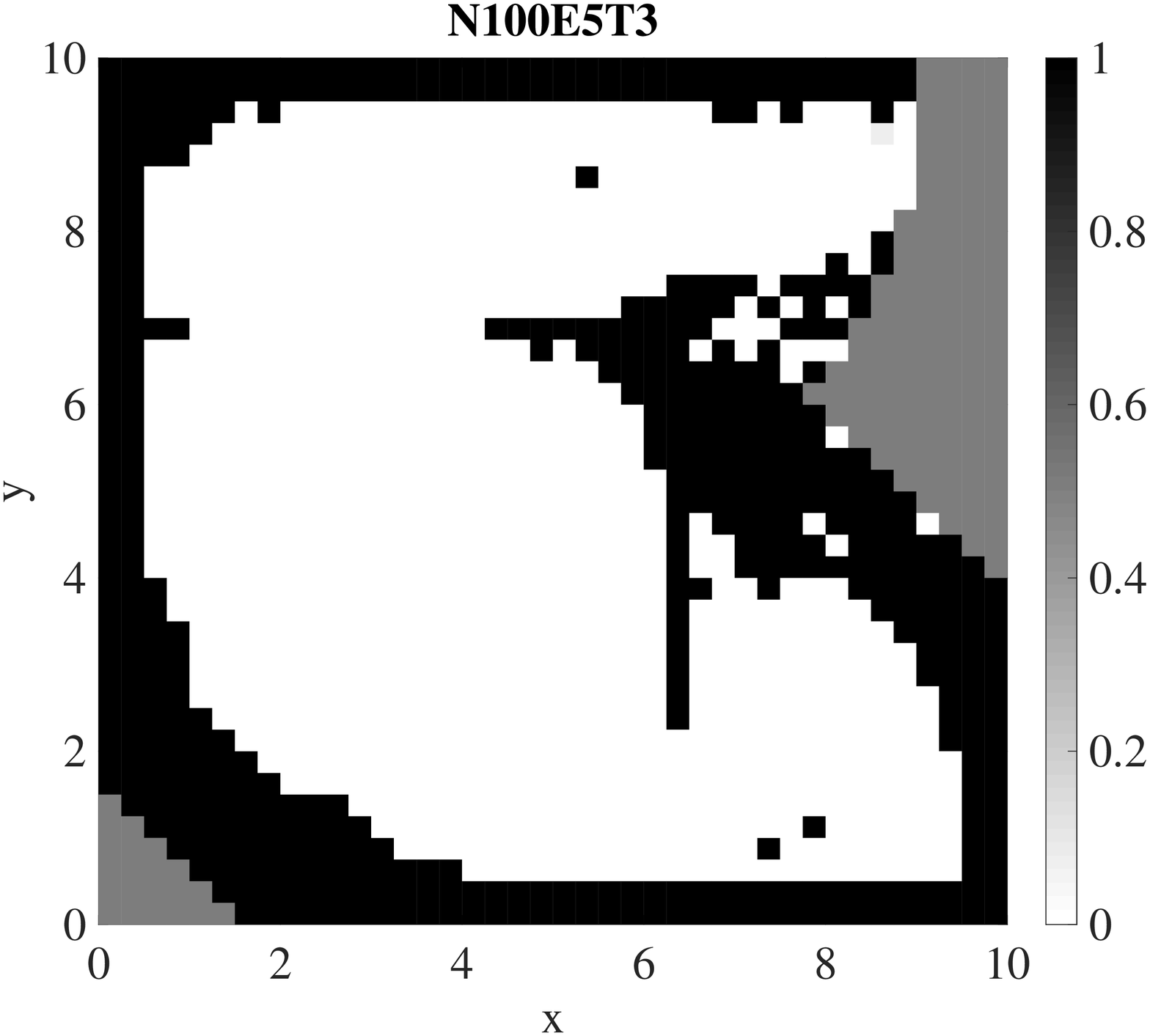}
}
\caption{Reference maps (top) and estimated probabilistic maps for three radar trajectories (namely, $\mathcal{T}_1$, $\mathcal{T}_2$ and $\mathcal{T}_3$) and for $N=16$ (middle) and $N=100$ (bottom).}
\label{fig:mapt}
\end{figure*}
In the sequel, we will indicate with ${m}_{i,1} \triangleq \left\{ m_{i}=1 \right\} $ the event for the $i$th cell of being occupied and with ${m}_{i,0} \triangleq \left\{ m_{i}=0 \right\}$ the opposite event. 
Recalling the statistical measurement model in \eqref{eq:stat_ebs}, we have
\begin{align}
    &p \left( \mathbf{e}_{k} \lvert {m}_{i,1}    \right) \propto \exp\left(- \sum_{b=1}^{\Nsteer} \sum_{s=1}^{\Nbin} \frac{\left(e_{bs, k} - {h}_{bs, k}\left( {m}_{i,1}  \right) \right)^2}{\text{var}\left(e_{bs, k}\lvert {m}_{i,1}  \right)} \right), \\
    &p \left( \mathbf{e}_{k} \lvert {m}_{i,0}     \right) \propto \exp\left(- \sum_{b=1}^{\Nsteer} \sum_{s=1}^{\Nbin} \frac{\left(e_{bs, k} - {h}_{bs, k}\left( {m}_{i,0}  \right) \right)^2}{\text{var}\left(e_{bs, k}\lvert {m}_{i,0}  \right)} \right),
\end{align}
\noindent where the measurement models $h_{bs, k}\left( m_{i,1} \right)$  and $h_{bs, k}\left( {m}_{i,0} \right)$  are computed using \eqref{eq:stat_ebs}  by only considering the contribution of the $i$th cell as
\begin{align}
    & {h}_{bs, k}\left( {m}_{i,1}  \right)\!\!=\!\! \begin{cases}\int_W \frac{L_\text{0}(f)\,\rho_{i}}{\left(d_{ik}\right)^4}  G^2(\theta_i-\theta_b, f)  \, df, \quad \, s=s_{i,k} \\
    0, \qquad\qquad\qquad\qquad\qquad\qquad\,\,  s\neq s_{i,k}
    \end{cases},
    \\
    &{h}_{bs, k}\left( {m}_{i,0}  \right)=0, \label{eq:model}
\end{align}
where $s_{i,k}= \Big \lfloor  \frac{2\,d_{ik}}{c\, \Ted} \Big\rfloor$ is the time index where the backscattered signal from the $i$th cell is expected to arrive.
Similarly, $\text{var}\left(e_{bs, k}\lvert m_{i,1} \right)$ and $\text{var}\left(e_{bs, k}\lvert {m}_{i,0} \right)$ can be found by injecting \eqref{eq:model} in \eqref{eq:stat_ebs}. 

In Fig.~\ref{fig:mapt}, we present examples of reconstructed maps, using different \ac{UAV} trajectories and number $N$ of antenna array elements. It is interesting to see how different trajectories result in different mapping accuracies  and, thus, one needs a trajectory optimization to enhance the final performance.

\paragraph{Target Detection}
We now describe the signal processing performed by the detector module of Fig.~\ref{fig:Stimatore}. This module determines if a target is present in the environment. 
We assume a scenario with unknown deterministic signals (i.e., those transmitted by the target) in \ac{AWGN} conditions, since multipath is neglected 
and obstacles can only obstruct the \ac{LOS} component.
In addition, we assume that the agent-\ac{UAV} has  knowledge of the target operating  bandwidth $W$.
After bandpass filtering over the bandwidth $W$, the received signal $r(t)$ is sampled at the Nyquist rate, thus obtaining the vector $\mathbf{y}=\left[y[1],\,\ldots ,\,y[n],\,\ldots ,\,y[N]\right]$, where $N=2\,T\,W$ is the number of samples,\footnote{In the rest of the manuscript, we assume $N \gg 1$.}  $T$ is the observation time window, and $N$ is an integer.
According to the aforementioned definitions, the normalized energy test statistic can be expressed by \cite{guidi2016joint} 
\begin{align}\label{eq:ynorm}
    \frac{2}{{N_0}_\nu} \int_{0}^{T}\left[ r(t)\right]^2 dt \simeq \frac{1}{\sigma_\nu^2} \sum_{n=0}^{N-1} \left|y[n]\right|^2\,,
\end{align}
where ${\sigma}_\nu^2 = {N_0}_{\nu}\,W$ represents the noise power and ${N_0}_\nu$ is the one-sided noise power spectral density of the receiver of the detector module.

Then, the related discrete time
detection problem can be written as
\begin{align}
&\mathcal{H}_0: \, y[n]=\nu[n],  \\
&\mathcal{H}_1: \, y[n]=x[n]+\nu[n],
\end{align}
where $x[n]$ and $\nu[n]$ are the $n$th samples of the low-pass representation of the signal and noise component, respectively, with $\nu[n]\sim \mathcal{N}(0,\,\sigma_\nu^2)$.

In our case, we consider the normalized energy test \cite{MarGioChi:J11,urkowitz1967energy} 
\begin{align}\label{eq:ENERGY}
    \Lambda_{\mathsf{ED}}\triangleq\frac{1}{\sigma_\nu^2} \cdot \sum_{n=1}^{N} \lvert y[n] \rvert^2 \underset{D_0}{\overset{D_1}{\gtrless}}   \xi,
\end{align}
which represents also the \ac{GLRT} when the signal that is being detected is unknown, as is the case here \cite{MarGioChi:J11,urkowitz1967energy}.
Notably, in absence of target, we have
\begin{align}\label{eq:ENERGY}
    \Lambda_{\mathsf{ED}}\simeq \frac{1}{\sigma_\nu^2} \cdot \sum_{n=1}^{N} \lvert \nu[n] \rvert^2\,,
\end{align}
which is distributed according to a central Chi-square distribution with $N$ degrees of freedom.
According to the considered model, the \ac{PFA} is given by \cite{MarGioChi:J11}
\begin{align}\label{eq:pfa}
 P_\text{FA}&=\frac{\Gamma\left( N, \frac{\xi}{2} \right)}{\Gamma\left(N \right)}= \tilde{\Gamma}\left({N},{\frac{\xi}{2}}\right),
\end{align} 
where $\Gamma(a,x)= \int_{x}^{\infty} x^{a-1}\, e^{-x} dx$ is the Gamma function, and $\tilde{\Gamma}\left( \cdot \right)$ is the Gamma regularized function.
According to the Neyman-Pearson criterion, we set the threshold according to a constraint on the desired $P_\text{FA}^\star$, so that it can be written by inverting \eqref{eq:pfa} in the form:
\begin{align}\label{eq:threshold}
    \xi=2\left[\text{Inv}\tilde{\Gamma}\left(N, P^\star_\text{FA}\right) \right] \,,
\end{align}
where $\text{Inv}\tilde{\Gamma}(\cdot, \cdot)$ is the inverse gamma regularized function.

According to the defined threshold, the probability of correct detection can be expressed as
\begin{align}\label{eq:pd}
    P_\text{D}=\mathcal{Q}_h (\sqrt{\lambda},\,\sqrt{\xi}),
\end{align}
with $\mathcal{Q}_h$ being the Marcum's $\mathcal{Q}$-function of order $h=N/2$, and $\lambda=\sum_{n} \left[x[n]\right]^2/\sigma_{\nu}^2$.
\begin{figure}[t]
\psfrag{x}[lc][lc][0.9]{FAR}
\psfrag{y}[lc][lc][0.9]{CDR}
\psfrag{0.5}[lc][lc][0.8]{$0.5$}
\psfrag{0.55}[lc][lc][0.8]{}
\psfrag{0.6}[lc][lc][0.8]{$0.6$}
\psfrag{0.65}[lc][lc][0.8]{}
\psfrag{0.7}[lc][lc][0.8]{$0.7$}
\psfrag{0.75}[lc][lc][0.8]{}
\psfrag{0.8}[lc][lc][0.8]{$0.8$}
\psfrag{0.85}[lc][lc][0.8]{}
\psfrag{0.9}[lc][lc][0.8]{$0.9$}
\psfrag{0.95}[lc][lc][0.8]{}
\psfrag{1}[lc][lc][0.8]{$1$
\psfrag{10}[lc][lc][0.8]{$10$}}
\psfrag{data1111111111}[lc][lc][0.7]{CDR, $d=10\,$m}
\psfrag{data11111111}[lc][lc][0.7]{$P_\text{D}$, $d=10\,$m}
\psfrag{data2}[lc][lc][0.7]{CDR, $d=13\,$m}
\psfrag{data22}[lc][lc][0.7]{$P_\text{D}$, $d=13\,$m}
\psfrag{data3}[lc][lc][0.7]{CDR, $d=15\,$m}
\psfrag{data33}[lc][lc][0.7]{$P_\text{D}$, $d=14\,$m}
\psfrag{data4}[lc][lc][0.7]{CDR, $d=17\,$m}
\psfrag{data44}[lc][lc][0.7]{$P_\text{D}$, $d=15\,$m}
\psfrag{data5}[lc][lc][0.7]{CDR, $d=17\,$m}
\psfrag{data55}[lc][lc][0.7]{$P_\text{D}$, $d=17\,$m}
\psfrag{data6}[lc][lc][0.7]{CDR, $d=20\,$m}
\psfrag{data66}[lc][lc][0.7]{$P_\text{D}$, $d=20\,$m}
\psfrag{data7}[lc][lc][0.7]{CDR, $d=25\,$m}
\psfrag{data77}[lc][lc][0.7]{$P_\text{D}$, $d=25\,$m}
\psfrag{data8}[lc][lc][0.7]{CDR, $d=30\,$m}
\psfrag{data88}[lc][lc][0.7]{$P_\text{D}$, $d=30\,$m}
\psfrag{data9}[lc][lc][0.7]{CDR, $d=35\,$m}
\psfrag{data99}[lc][lc][0.7]{$P_\text{D}$, $d=35\,$m}
\centerline{
\includegraphics[width=0.95\linewidth,draft=false]{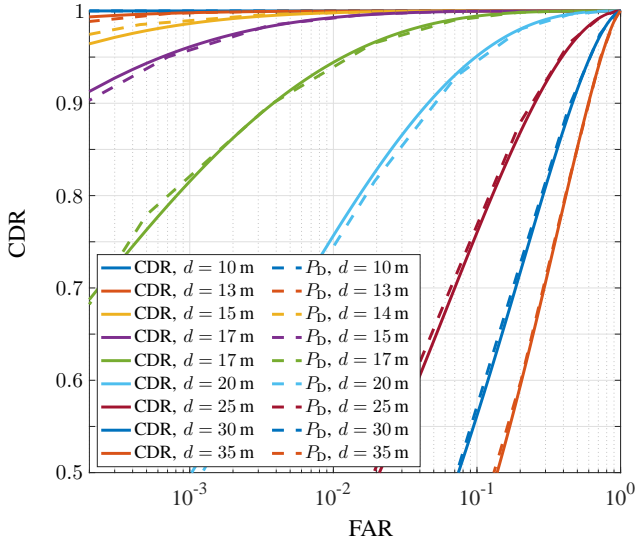} 
}
\caption{Continuous line: simulated \ac{CDR} vs. \ac{FAR}. Dashed line: theoretical results.} 
\label{fig:rocsim}
\end{figure}
\begin{algorithm}[t!]
\SetAlgoLined
\textbf{Parameters}: Set the discount rate $\gamma$; \\
Set the number of future instants (i.e., the horizon) $T_{\mathsf{H}}$; \\
Set the mission time $\Tmission$; \\
Set the probability of taking a random action $\epsilon$;   \\
\textbf{Initialization}:
Initialize the map estimate $\hat{\mathbf{m}}_0$; \\
Set the initial UAV position $\mathbf{s}_{\mathsf{U},0}$; \\
Initialize the state $\hat{\mathbf{s}}_{k=0}=\left[ \mathbf{s}_{\mathsf{U},0}, \hat{\mathsf{t}}_0, \hat{\mathbf{m}}_0 \right]$ 
; \\
  \For{time instant $k<\Tmission$}{
  Set $\epsilon$ as a function of $\Tmission$, according to \eqref{eq:eps};\\
  Generate a random value $\epsilon_k$;\\
  \eIf{ $\epsilon_k<\epsilon$}{
   Choose a random action $\mathbf{a}_k \in \mathcal{A}$\;
   }{
   For each possible $\pi$, evaluate $Q_\pi$ according to the expected rewards $r\left({\mathbf{s}}_k, \mathbf{a}_k  \right) \lvert_{{\mathbf{s}}_k=\hat{\mathbf{s}}_k}$ until $T_\mathsf{H}-1$;\\
   Choose an action $\mathbf{a}_k \in \mathcal{A}$ according to \eqref{eq:pistar};
  }
  Agent moves to the new state, , $\mathbf{s}_{\mathsf{U},k+1}= \mathbf{s}_{\mathsf{U},k}+\mathbf{a}_{k}$;\\
  Agent acquires measurements (i.e., $\mathbf{e}_k$ and $\mathbf{y}_k$); \\
  Agent performs state estimation, $\hat{\mathbf{s}}_{k+1}$, using \eqref{eq:ML} and \eqref{eq:ENERGY};
 }
 \caption{``Policy Estimator" for inferring $\pi^*$ \cite{sutton2018reinforcement}}
 \label{alg:qalg}
\end{algorithm}
\subsection{Actions}
In this problem, the actions are the control signals to be applied by the \ac{UAV} for moving from one cell to another, i.e., $\ak=\Deltapuk$.
Since we consider a stationary environment, i.e. $m_{i, k+1}=m_{i,k}$, $\mathsf{t}_{ k+1}=\mathsf{t}_{k}$ $\forall \, i,k$, here we focus only on the \ac{UAV} transition dynamics, that is $\puavf  =\puavk +\Deltapuk$.
\subsection{Rewards}
The expected reward $r\left(\mathbf{s}_k, \mathbf{a}_k \right)$ is computed as 
\begin{align}\label{eq:reward}
    r\left(\mathbf{s}_k, \mathbf{a}_k \right)
    &=
    \mathbb{E}\left[R_{k+1}=r_{k+1} \lvert \mathbf{s}_k, \mathbf{a}_k \right] \nonumber \\
    &=\sum_{r_{k+1} \in \rewardspace}\, r_{k+1}\, \sum_{\mathbf{s}_{k+1} \in \statespace} p\left( \mathbf{s}_{k+1} \lvert \mathbf{s}_{k}, \mathbf{a}_{k}\right).
 \end{align}

In our decision and mapping problem, we define the following rewards:
\begin{itemize}
    \item $\mathrm{r}_{\mathrm{map}}$: Mapping reward. To obtain a quantitative evaluation of the mapping performance, we use the entropy $H$ of the map according to \cite{mafi2011information}
\begin{equation}\label{eq:imsim}
\mathrm{r}_{\mathrm{map},k+1}=
  \bigg[H_{k+1 \lvert k}(\mathbf{m})\bigg]^{-1}\,,
\end{equation}
where  
\begin{align}
H_{k+1 \lvert k}(\mathbf{m})=&\sum_{i \in \mathcal{I}}  -b_{k+1 \lvert k}({m}_{i,1})  \, \log\left( b_{k+1 \lvert k}({m}_{i,1})  \right)+ \nonumber \\
&-b_{k+1 \lvert k}({m}_{i,0})  \, \log\left( b_{k+1 \lvert k}({m}_{i,0})  \right),
\end{align}
represents the entropy of the estimated map from the acquired measurements by time $k$. 
\item $\mathrm{r}_{\mathrm{d}}$: Detection reward.  For each cell, we have
\begin{equation}\label{eq:pdetection}
    \mathrm{r}_{\mathrm{d}, k+1} = 1- P_{\mathrm{e}, k+1} 
\end{equation}
where $P_{\mathrm{e}, k+1}$ is defined in \eqref{eq:pe} and is computed for each \ac{UAV} position, where $\Ppa=P_\text{FA}$ and $\Pap=P_\text{M}=1-P_\text{D}$ are the false alarm and missed detection probabilities, with
  $P_\text{FA}$ and $P_\text{D}$  defined in \eqref{eq:pfa} and \eqref{eq:pd}, and where $P_1=P\left( \Hp \right)$ and $P_0=P\left( \Ha \right)$ are the probabilities of a target in the environment.
In our work, we suppose to have the target always present in the environment, so that we have $\Pa=0$ and $\Pp=1$. Consequently, \eqref{eq:pdetection} becomes $\mathrm{r}_{\mathrm{d}, k+1} =   P_{\text{D}, k+1}$.
\end{itemize}

\section{Numerical Results}
\label{sec:results}
\subsection{Non Optimized Trajectory - Mapping Results}\label{sec:nonauto}

For the considered case study, we accounted for an {EIRP} of $5\,$dBm, a receiver noise figure of $4\,$dB, and a transmitted pulse with bandwidth of $1\,$GHz centered at $\fcradar=60\,$GHz. The number of pulses depends on the fixed scanning time, i.e.,  $\Np=\lceil T_{\mathrm{obs}} / T_{\mathrm{f}} \rceil$ where  $T_{\mathrm{obs}}=T_{\mathrm{scan}}/\Nsteer$ is the observation time, and $T_{\mathrm{scan}}=80\,\mu$s is the time needed to perform an entire scan operation. For example, for $N=4\times4$ ($N=10\times10$), the steering directions were set to $\Nsteer=8$ ($\Nsteer=20$) leading to a number of pulses $\Np=334$ ($\Np=134$).

\begin{figure*}[t]
\psfrag{x}[c][c][1]{$x$ [m]}
\psfrag{y}[c][c][1]{$y$ [m]}
\psfrag{t}[rc][rc][1]{target}
\psfrag{p}[c][c][0.8]{$\mathbf{p}_0$}
\psfrag{p2}[c][c][0.8]{$\mathbf{p}_{\Tmission}$}
\centerline{
\includegraphics[width=0.45\linewidth,draft=false]{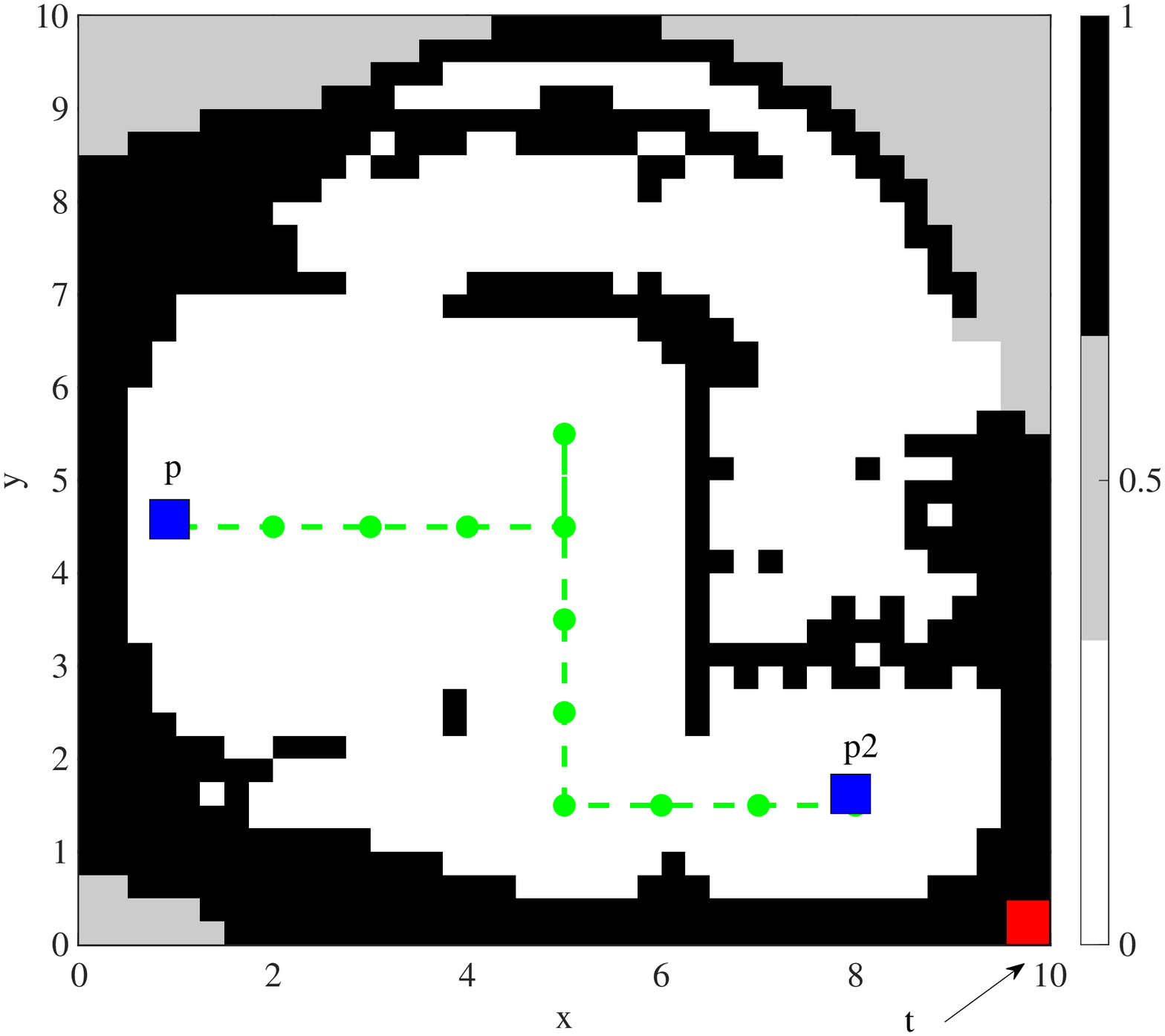}
\quad \quad
\includegraphics[width=0.45\linewidth,draft=false]{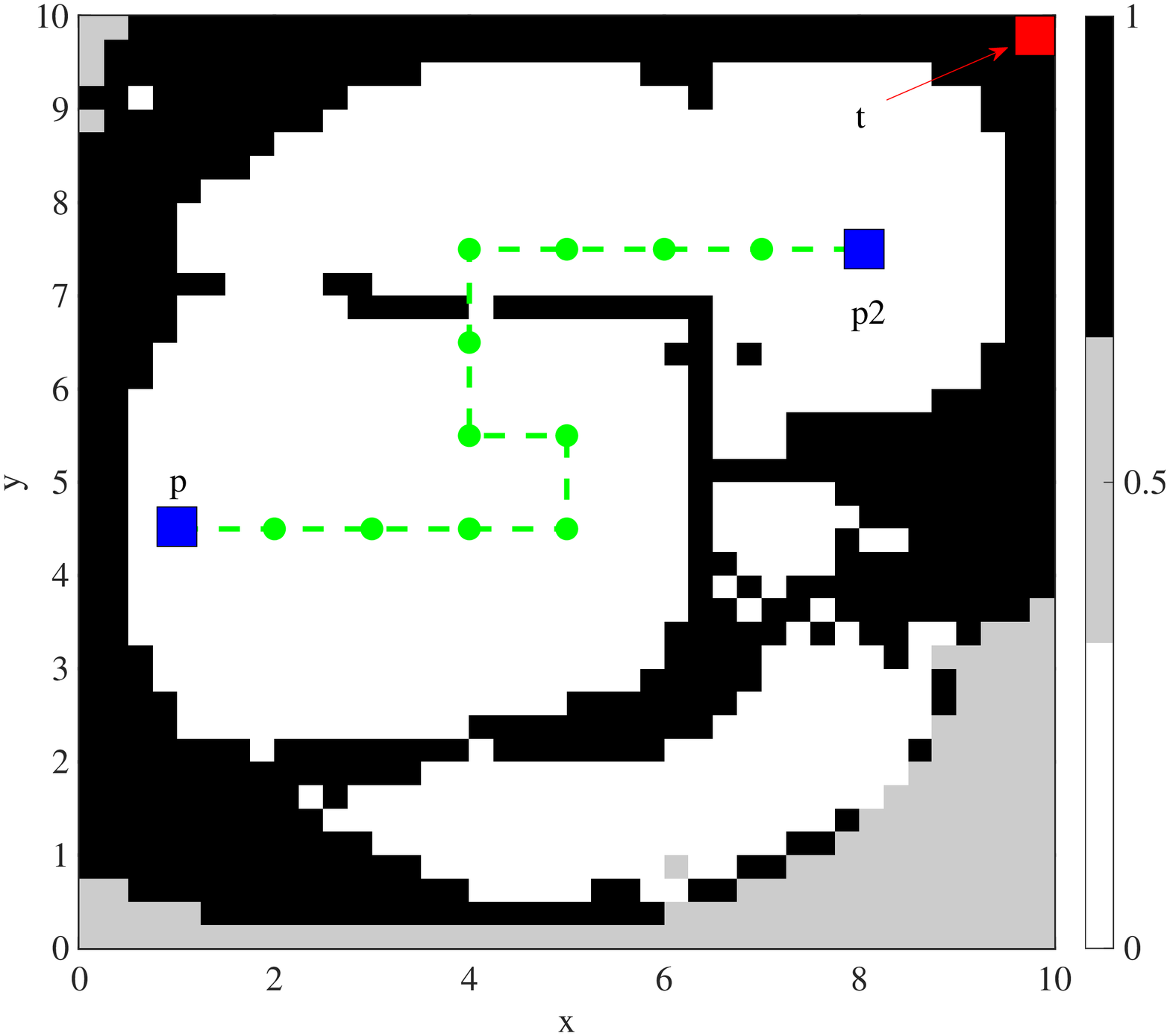} 
}
\vspace{0.3cm}
\centerline{
\includegraphics[width=0.45\linewidth,draft=false]{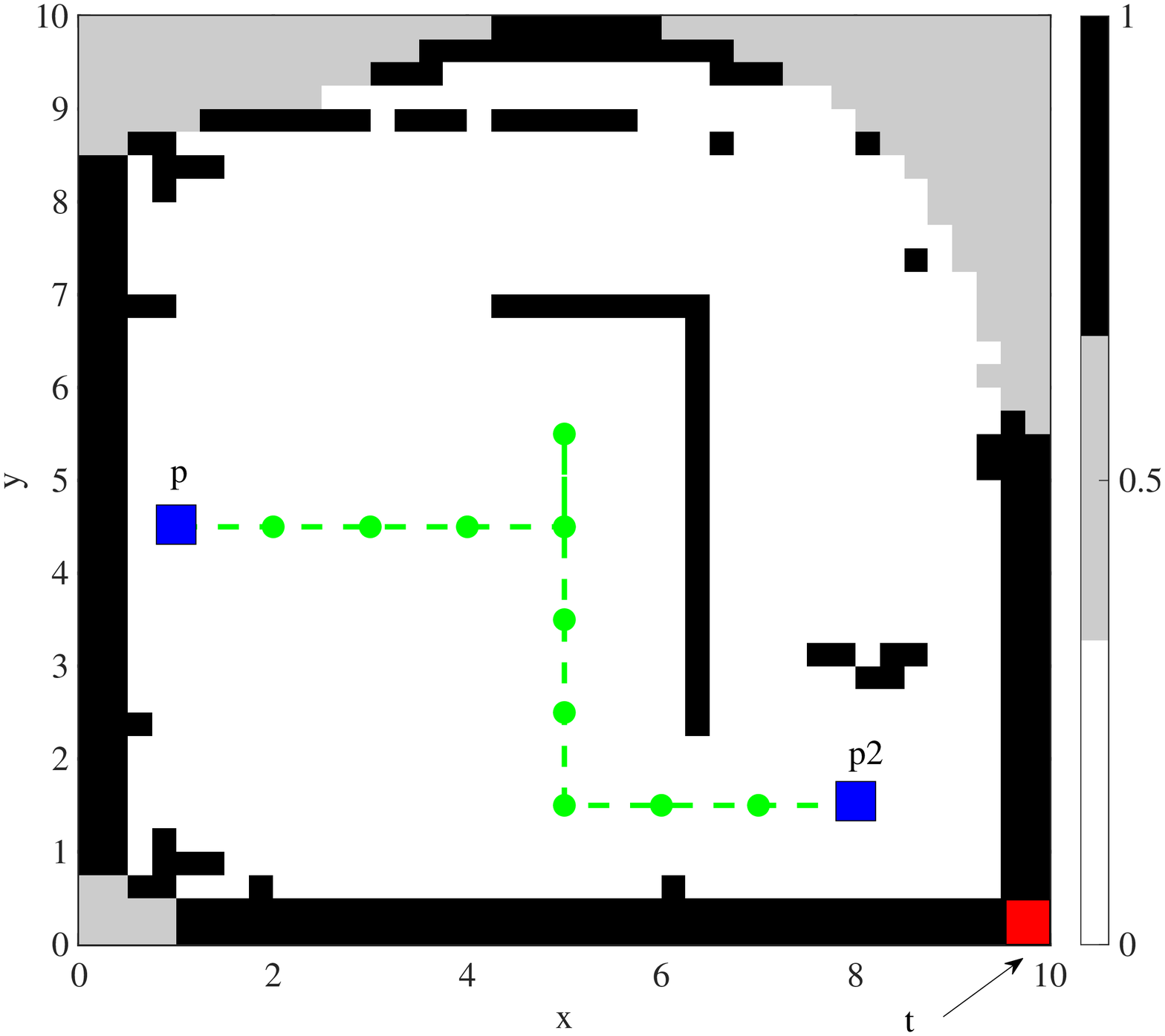} 
\quad \quad 
\includegraphics[width=0.45\linewidth,draft=false]{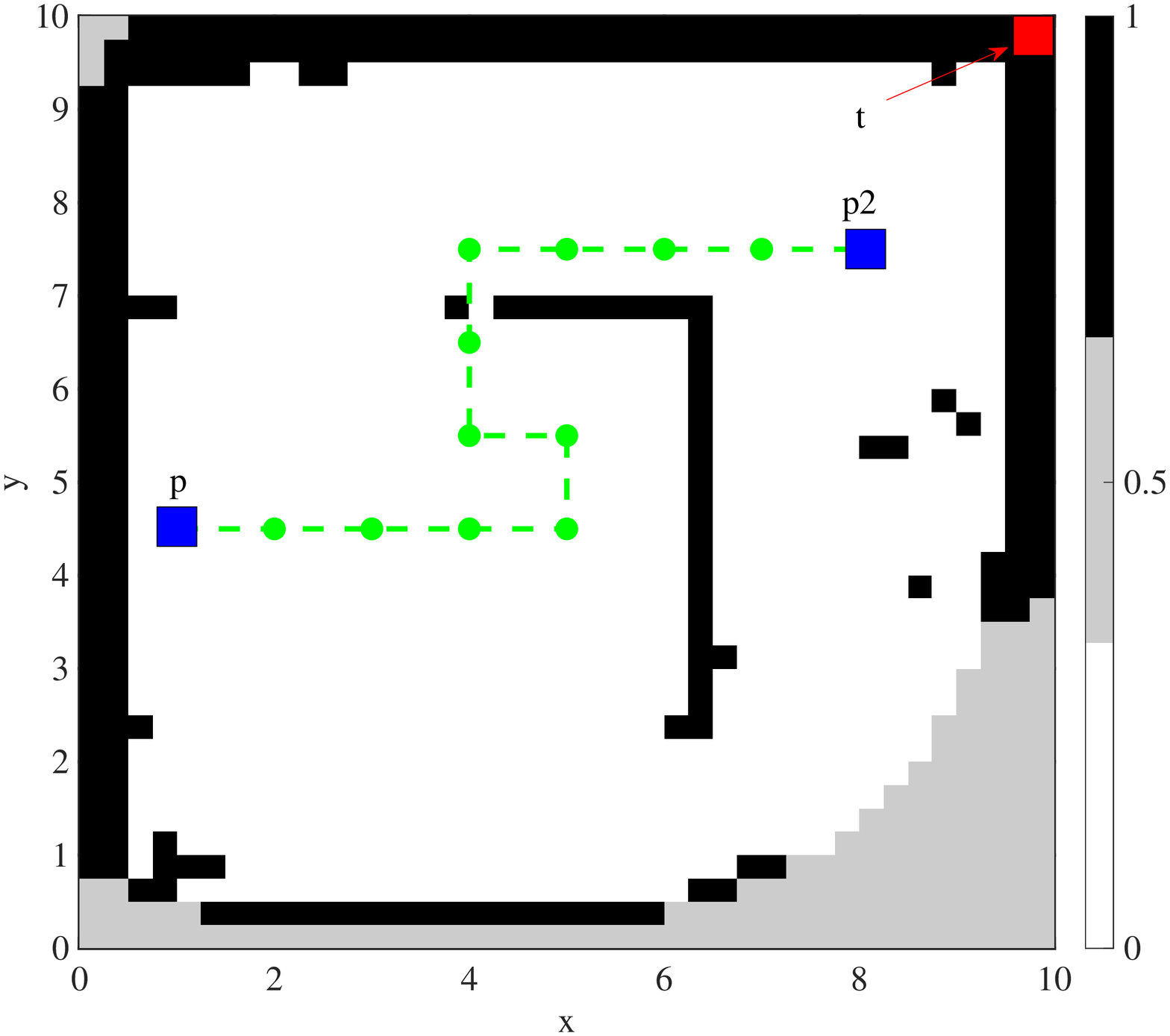} 
}
\caption{UAV autonomous navigation for $N=16$ (top) and $N=100$ (bottom). Left: target placed on the bottom-right corner. Right: target placed on the bottom-left corner.}
\label{fig:scenarioRL_uav_3}
\end{figure*}
Figure~\ref{fig:mapt} presents the outcomes of the ``\textit{State Estimator}" block dedicated to the map reconstruction, i.e., the estimated map $\hat{\mathbf{m}}_k$. 
\begin{figure*}[t]
\psfrag{x}[c][c][1]{$x$ [m]}
\psfrag{y}[c][c][1]{$y$ [m]}
\psfrag{t}[rc][rc][1]{target}
\psfrag{p}[c][c][0.8]{$\mathbf{p}_0$}
\psfrag{p2}[c][c][0.8]{$\mathbf{p}_{\Tmission}$}
\centerline{
\includegraphics[width=0.45\linewidth,draft=false]{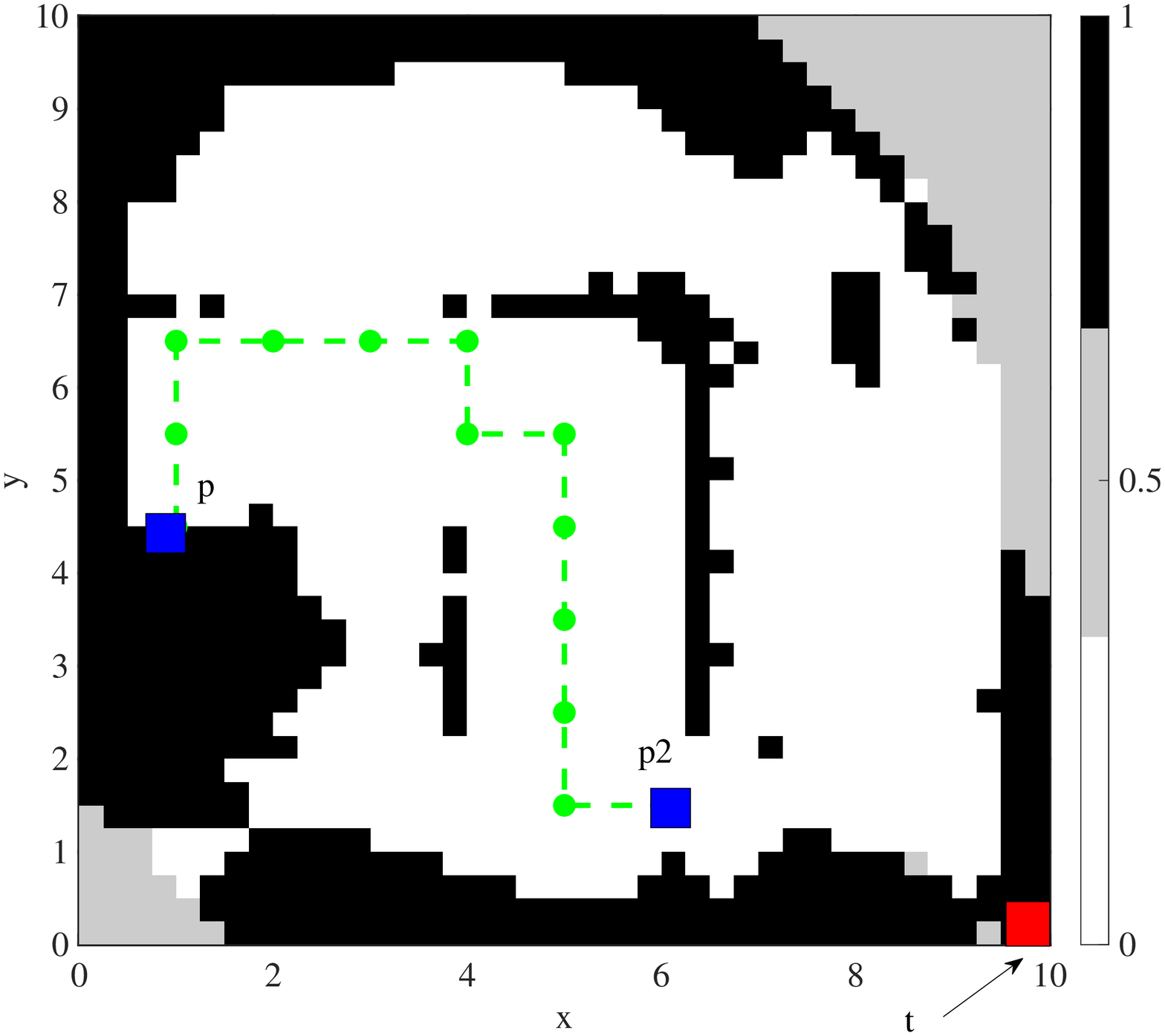} 
\quad \quad 
\includegraphics[width=0.45\linewidth,draft=false]{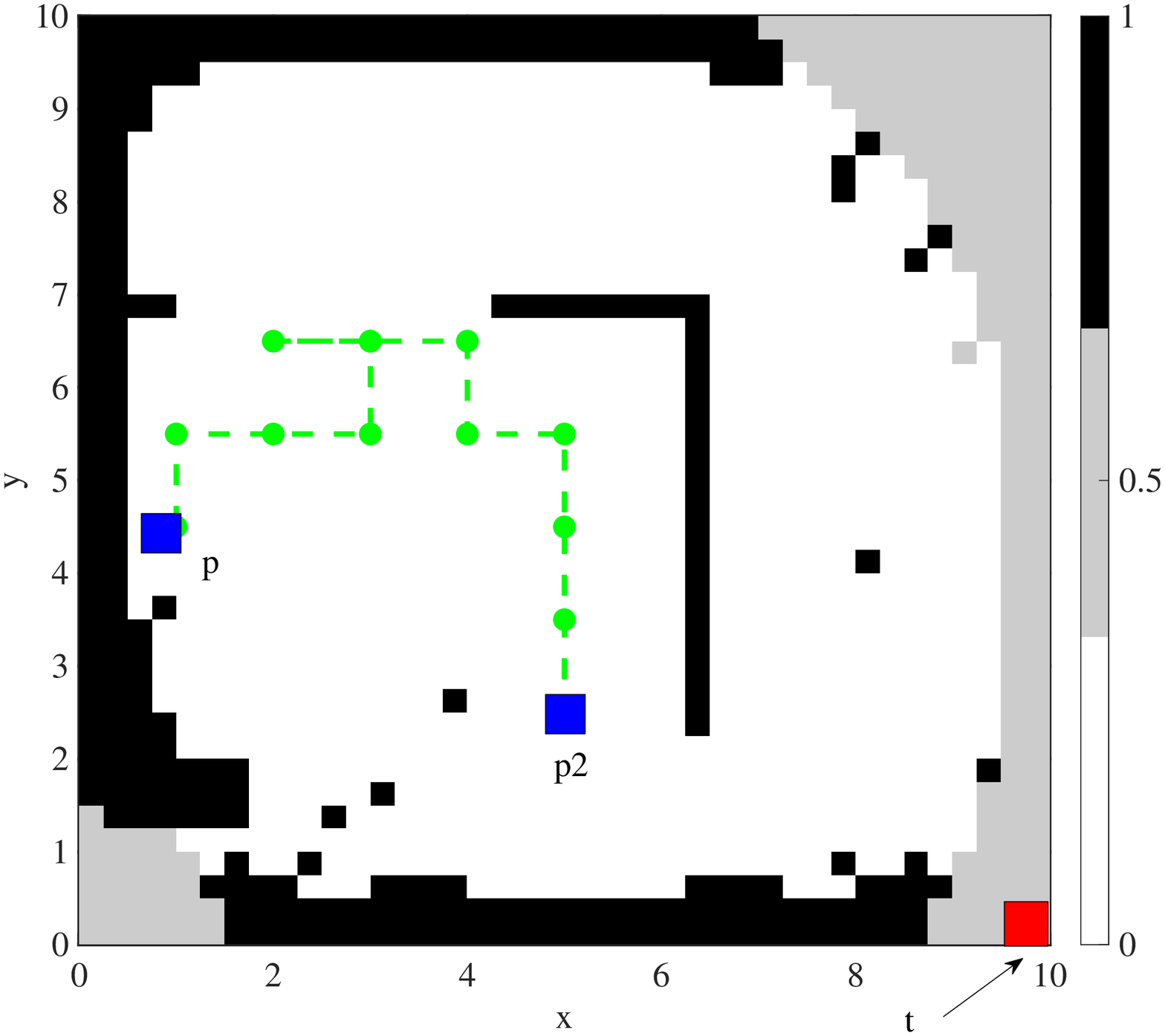}
}
\caption{UAV autonomous navigation for $N=16$ (left) and $N=100$ (right) and with a target placed on the bottom-right corner. The exploration rate was set according to the mission time as in \eqref{eq:eps}.}
\label{fig:new_res}
\end{figure*}
The reference maps are shown in Fig.~\ref{fig:mapt}-top for different \ac{UAV} trajectories and mission times, namely $\mathcal{T}_1$-$\mathcal{T}_2$-$\mathcal{T}_3$, whereas the estimated maps are in Fig.~\ref{fig:mapt}-middle and Fig.~\ref{fig:mapt}-bottom for $N=4\times4$ and $N=10\times10$, respectively. The color scale indicates the probability of occupancy, for example the black cells are occupied with probability equal to $1$, whereas the white cells are free. Initially, we suppose to have a complete uncertainty about the map, i.e., $b_0\left(m_{i,1} \right)=b_0\left(m_{i,0} \right)=0.5$, $\forall i$. The radar trajectory is depicted with green dots. 

As expected, an increased number of antennas, leading to an increased antenna gain and angular resolution, results in a better map reconstruction.
Moreover, it is clearly visible that the mapping performance depends on the \ac{UAV} trajectories and observation time (i.e., on the number of measurement points). Consequently, an optimization of the radar trajectory becomes essential, especially for emergency situations.

\subsection{Target Detection Design}
Figure~\ref{fig:rocsim} shows the detection results in terms of \ac{ROC} that allows the assessment of the target detection performance as a function of the intended \ac{PFA} (i.e., of the threshold) and \ac{SNR}.

More specifically, in Fig.~\ref{fig:rocsim}, we show results of simulations when a target is at different distances from the receiver, $d=\left[10,\, 13,\, 14,\, 15,\, 17,\, 20,\, 25,\, 30,\, 35\right]\,$m. Consequently, the non-central parameter $\lambda$ changes accordingly. 
By running the energy test in \eqref{eq:ENERGY}, we have computed the \ac{CDR} and \ac{FAR} as
\begin{align}
    \mathsf{CDR}&= \frac{1}{N_{\mathsf{1}}} \sum_{i=1}^{N_{\mathsf{1}}} \mathbf{1}\left( \Lambda_{\mathsf{ED}} \overset{D_1}{>}  \xi \lvert \Hp \right),   
\end{align}
\begin{align}
    \mathsf{FAR}&= \frac{1}{N_{\mathsf{0}}} \sum_{i=1}^{N_{\mathsf{0}}} \mathbf{1}\left( \Lambda_{\mathsf{ED}} \overset{D_1}{>}  \xi \lvert \Ha \right),
\end{align}
where $N_{\mathsf{1}}$ and $N_{\mathsf{0}}$ are the number of times that the target was present and absent in the simulations, with  $N_{\mathsf{MC}}=N_{\mathsf{1}}+N_{\mathsf{0}}$ being the overall number of iterations, and $\mathbf{1}(x)$ is a function that equals to one when its argument is true, and $0$ otherwise. The results are compared with theoretical curves of probabilities obtained by  \eqref{eq:pfa}-\eqref{eq:pd}.

The target detector performance is used in the sequel to model the detection reward for each action. 

\subsection{Optimized Trajectory - Mapping \& Detection Results}

We now describe the  \ac{UAV} autonomous navigation in order to optimize target detection and environment mapping. If not otherwise indicated, the same parameters of Sec.~\ref{sec:nonauto} were considered.

As a first approximation, we consider the agent equipped with a proximity radar for detection of the presence of close obstacles, so that collisions are avoided. The implemented ``\textit{Policy Estimator}" is outlined in Algorithm 1 and is inspired by a Q-learning approach, where we set $\gamma=0.89$, and $T_{\mathsf{H}}=4$ step. The state was initialized considering $\hat{m}_i=0.5$, $\forall i$, and $\mathsf{t}=1$.
The mission time was fixed to $\Tmission=13\,$step, and
the radar could perform only lateral and vertical movements of $1\,$m. The \ac{UAV} velocity was set to $1\,$m/step.

For each decision taken at time instant $k$, the agent accounted for all the possible state combinations up to time instant $k+T_{\mathsf{H}}$.
The desired \ac{PFA} was set to $P_{\mathsf{FA}}^\star=10^{-3}$.

Figure~\ref{fig:scenarioRL_uav_3} shows four examples, obtained for $N=16$ (top) and $N=100$ (bottom). 
In particular, the scenarios on the left present a target located in $x=10,\,y=0$, whereas the two on the right consider a target located in $x=10,\,y=10$. The \ac{UAV} is initially located cyan at  $\mathbf{s}_{\mathsf{U},0}=\left[ 1,\,4.5 \right]$. Interestingly, the agent is capable of reconstructing the environment while moving towards cells closer to the target that exhibit a higher $P_\text{D}$, and thus are more advantageous in terms of joint rewards. Even if several cells in the map are still uncertain (i.e., with final belief of $0.5$), for $N=100$, the \ac{UAV} can reconstruct the environment reliably.

So far, we have investigated the following two scenarios: (i) non-optimized trajectories (see Fig.~\ref{fig:mapt}-top); (ii) optimized trajectories for tackling the goal of the mission (see Fig.~\ref{fig:scenarioRL_uav_3}). Even if in the second scenario the agent prioritizes the tasks of the mission, some parts of the map may never be explored.
To mitigate such a detrimental effect, we included also a rate of exploration  $\epsilon$ dependent on the mission time $\Tmission$ as 
\begin{align}\label{eq:eps}
    \epsilon=\begin{cases}
    0.8 \quad \text{if} \quad k<\Tmission/4 \\
    0.4 \quad \text{if} \quad \Tmission/4/\leq k<\Tmission/2 \\ 
    0 \quad \quad\!\! \text{Otherwise} \,.
    \end{cases}
\end{align}
By this rule, initially, the agent chooses the actions randomly, whereas for $k>\Tmission/2$, it evaluates the expected awards and chooses the action accordingly.  

In this respect, Fig.~\ref{fig:new_res} shows the obtained curves for $N=16$ (left) and $N=100$ (right). In both cases,  the final position of the agent is not as optimized for target detection (i.e., it is more distant from the target) with respect to their counterparts in Fig.~\ref{fig:scenarioRL_uav_3}-bottom, but the initial randomness allows for better exploration of the environment. 

Such considerations leave an open door for further research of this issue.
For example,  a more accurate design of $\epsilon$ can be performed according to the available $\Tmission$. Another solution could be based on the inclusion of an ad-hoc reward weight that accounts for the coverage of the map. This last approach is very promising but requires that different types of rewards are suitably weighted in order to form a global reward.  

Finally, in the perspective of implementing collaborative approaches for multiple agents, the Q-values can be diffused among cooperative agents to enhance the learning mechanism. 
\section{Concluding Remarks} 
\label{sec:conclusions}
In this paper, we have investigated the possibility of empowering \acp{UAV} with \ac{RL} capabilities for trajectory optimization. The applications of interest might be an emergency or safety scenario where a single \ac{UAV}  reconstructs a map of an unknown environment while minimizing the error of detecting the presence of a target.
In contrast with non-optimized trajectory design (e.g., based on fixed waypoint paths), the results obtained using \ac{RL} are promising since the agent exhibits interesting capabilities in choosing the trajectory while achieving reliable performance in terms of detection and mapping accuracy in a given mission time.
Future work will consider a multiple target scenario, a performance comparison in terms of mission time, and the adoption of a \ac{POMDP} for taking into consideration the state estimation uncertainty in the decision making process. 


\section*{Acknowledgments}
{This work has received funding from the European Union's Horizon 2020 research and innovation programme under the Marie Sklodowska-Curie project AirSens (grant no. 793581) and from the PRIMELOC project funded by EU H2020.}

\bibliographystyle{IEEEtran}

\begin{thebibliography}{10}
\providecommand{\url}[1]{#1}
\csname url@samestyle\endcsname
\providecommand{\newblock}{\relax}
\providecommand{\bibinfo}[2]{#2}
\providecommand{\BIBentrySTDinterwordspacing}{\spaceskip=0pt\relax}
\providecommand{\BIBentryALTinterwordstretchfactor}{4}
\providecommand{\BIBentryALTinterwordspacing}{\spaceskip=\fontdimen2\font plus
\BIBentryALTinterwordstretchfactor\fontdimen3\font minus
  \fontdimen4\font\relax}
\providecommand{\BIBforeignlanguage}[2]{{%
\expandafter\ifx\csname l@#1\endcsname\relax
\typeout{** WARNING: IEEEtran.bst: No hyphenation pattern has been}%
\typeout{** loaded for the language `#1'. Using the pattern for}%
\typeout{** the default language instead.}%
\else
\language=\csname l@#1\endcsname
\fi
#2}}
\providecommand{\BIBdecl}{\relax}
\BIBdecl

\bibitem{marconi2012sherpa}
L.~Marconi \emph{et~al.}, ``The {SHERPA} project: Smart collaboration between
  humans and ground-aerial robots for improving rescuing activities in alpine
  environments,'' in \emph{Proc. IEEE Int. Symp. Safety, Security, Rescue
  Robot.}, 2012, pp. 1--4.

\bibitem{merwaday2016improved}
A.~Merwaday \emph{et~al.}, ``Improved throughput coverage in natural disasters:
  Unmanned aerial base stations for public-safety communications,'' \emph{IEEE
  Veh. Technol. Mag.}, vol.~11, no.~4, pp. 53--60, 2016.

\bibitem{wang2019autonomous}
C.~Wang \emph{et~al.}, ``Autonomous navigation of {UAV}s in large-scale complex
  environments: A deep reinforcement learning approach,'' \emph{IEEE Trans.
  Veh. Technol.}, vol.~68, no.~3, pp. 2124--2136, 2019.

\bibitem{hugler2018radar}
P.~H{\"u}gler \emph{et~al.}, ``Radar taking off: {N}ew capabilities for
  {UAV}s,'' \emph{IEEE Microw. Mag.}, vol.~19, no.~7, pp. 43--53, 2018.

\bibitem{guvenc2018detection}
I.~Guvenc \emph{et~al.}, ``Detection, tracking, and interdiction for amateur
  drones,'' \emph{IEEE Commun. Mag.}, vol.~56, no.~4, pp. 75--81, 2018.

\bibitem{hugler201877}
P.~H{\"u}gler, M.~Geiger, and C.~Waldschmidt, ``77 {GHz} radar-based altimeter
  for unmanned aerial vehicles,'' in \emph{Proc. IEEE Radio Wireless Symp.},
  2018, pp. 129--132.

\bibitem{BenJam:C11}
P.~{Benavidez} and M.~{Jamshidi}, ``Mobile robot navigation and target tracking
  system,'' in \emph{Proc. 6th Int. Conf. Sys. of Sys. Eng.}, 2011, pp.
  299--304.

\bibitem{mohammed2014uavs}
F.~Mohammed \emph{et~al.}, ``{UAV}s for smart cities: Opportunities and
  challenges,'' in \emph{Proc. Int. Conf. Unmanned Aircraft Sys.}, 2014, pp.
  267--273.

\bibitem{guerra2020dynamic}
A.~Guerra, D.~Dardari, and P.~M. Djuric, ``Dynamic radar network of {UAV}s: A
  joint navigation and tracking approach,'' \emph{arXiv preprint
  arXiv:2001.04560}, 2020.

\bibitem{guerra2019dynamic}
------, ``Dynamic radar networks of {UAV}s,'' \emph{IEEE Veh. Tech. Mag.},
  2020.

\bibitem{sutton2018reinforcement}
R.~S. Sutton and A.~G. Barto, \emph{Reinforcement learning: An
  introduction}.\hskip 1em plus 0.5em minus 0.4em\relax MIT press, 2018.

\bibitem{kuss2004gaussian}
M.~Kuss and C.~E. Rasmussen, ``Gaussian processes in reinforcement learning,''
  in \emph{Advances in neural information processing systems}, 2004, pp.
  751--758.

\bibitem{ciftler2017indoor}
B.~S. Ciftler, A.~Tuncer, and I.~Guvenc, ``Indoor {UAV} navigation to a
  {R}ayleigh fading source using {Q}-learning,'' \emph{arXiv preprint
  arXiv:1705.10375}, 2017.

\bibitem{wang2018reinforcement}
L.~Wang \emph{et~al.}, ``Reinforcement learning-based waveform optimization for
  mimo multi-target detection,'' in \emph{Proc. IEEE 52nd Asilomar Conf.
  Signals, Sys., Comput.}, 2018, pp. 1329--1333.

\bibitem{jiang2019end}
W.~Jiang, A.~M. Haimovich, and O.~Simeone, ``End-to-end learning of waveform
  generation and detection for radar systems,'' \emph{arXiv preprint
  arXiv:1912.00802}, 2019.

\bibitem{malhotra1997learning}
R.~Malhotra, E.~P. Blasch, and J.~D. Johnson, ``Learning sensor-detection
  policies,'' in \emph{Proc. IEEE National Aerosp. Electron. Conf.}, vol.~2,
  1997, pp. 769--776.

\bibitem{selvi2018use}
E.~Selvi \emph{et~al.}, ``On the use of {M}arkov decision processes in
  cognitive radar: An application to target tracking,'' in \emph{Proc. IEEE
  Radar Conf.}, 2018, pp. 0537--0542.

\bibitem{pham2018autonomous}
H.~X. Pham \emph{et~al.}, ``Autonomous uav navigation using reinforcement
  learning,'' \emph{arXiv preprint arXiv:1801.05086}, 2018.

\bibitem{liu2019reinforcement}
X.~Liu, Y.~Liu, and Y.~Chen, ``Reinforcement learning in multiple-uav networks:
  Deployment and movement design,'' \emph{IEEE Trans. Veh. Technol.}, vol.~68,
  no.~8, pp. 8036--8049, 2019.

\bibitem{saxena2019optimal}
V.~Saxena, J.~Jald{\'e}n, and H.~Klessig, ``Optimal {UAV} base station
  trajectories using flow-level models for reinforcement learning,'' \emph{IEEE
  Trans. Cogn. Commun. Netw.}, 2019.

\bibitem{bayerlein2018trajectory}
H.~Bayerlein, P.~De~Kerret, and D.~Gesbert, ``Trajectory optimization for
  autonomous flying base station via reinforcement learning,'' in \emph{Proc.
  IEEE 19th Int. Workshop Signal Process. Adv. Wireless Commun.}, 2018, pp.
  1--5.

\bibitem{theile2020uav}
M.~Theile \emph{et~al.}, ``{UAV} coverage path planning under varying power
  constraints using deep reinforcement learning,'' \emph{arXiv preprint
  arXiv:2003.02609}, 2020.

\bibitem{ragi2013uav}
S.~Ragi and E.~K. Chong, ``{UAV} path planning in a dynamic environment via
  partially observable markov decision process,'' \emph{IEEE Trans. Aerosp.
  Electron. Sys.}, vol.~49, no.~4, pp. 2397--2412, 2013.

\bibitem{kaelbling1998planning}
L.~P. Kaelbling, M.~L. Littman, and A.~R. Cassandra, ``Planning and acting in
  partially observable stochastic domains,'' \emph{Artificial intelligence},
  vol. 101, no. 1-2, pp. 99--134, 1998.

\bibitem{thrun2000monte}
S.~Thrun, ``Monte carlo {POMDP}s,'' in \emph{Adv. Neural Info. Process. Sys.},
  2000, pp. 1064--1070.

\bibitem{GueGuiDar:J18}
A.~{Guerra}, F.~{Guidi}, and D.~{Dardari}, ``Single-anchor localization and
  orientation performance limits using massive arrays: {MIMO} vs.
  beamforming,'' \emph{IEEE Trans. Wireless Commun.}, vol.~17, no.~8, pp.
  5241--5255, 2018.

\bibitem{ShaEtAl:J18}
A.~{Shahmansoori} \emph{et~al.}, ``Position and orientation estimation through
  millimeter-wave {MIMO} in 5{G} systems,'' \emph{IEEE Trans. Wireless
  Commun.}, vol.~17, no.~3, pp. 1822--1835, 2018.

\bibitem{vukmirovic2019direct}
N.~Vukmirovi{\'c} \emph{et~al.}, ``Direct wideband coherent localization by
  distributed antenna arrays,'' \emph{Sensors}, vol.~19, no.~20, p. 4582, 2019.

\bibitem{GuiGueDar:J16}
F.~Guidi, A.~Guerra, and D.~Dardari, ``Personal mobile radars with
  millimeter-wave massive arrays for indoor mapping,'' \emph{IEEE Trans. Mobile
  Comput.}, vol.~15, no.~6, pp. 1471--1484, Jun. 2016.

\bibitem{JouEtAl:J17}
C.~{Jouanlanne} \emph{et~al.}, ``Wideband linearly polarized transmitarray
  antenna for 60 {GH}z backhauling,'' \emph{IEEE Trans. Antennas Propag.},
  vol.~65, no.~3, pp. 1440--1445, 2017.

\bibitem{ghosh2019inclusive}
S.~Ghosh and D.~Sen, ``An inclusive survey on array antenna design for
  millimeter-wave communications,'' \emph{IEEE Access}, vol.~7, pp.
  83\,137--83\,161, 2019.

\bibitem{guerra2018occupancy}
A.~Guerra \emph{et~al.}, ``Occupancy grid mapping for personal radar
  applications,'' in \emph{Proc. Stat. Signal Process. Workshop}, 2018, pp.
  766--770.

\bibitem{urkowitz1967energy}
H.~Urkowitz, ``Energy detection of unknown deterministic signals,'' \emph{Proc.
  IEEE}, vol.~55, no.~4, pp. 523--531, 1967.

\bibitem{thrun2003learning}
S.~Thrun, ``Learning occupancy grid maps with forward sensor models,''
  \emph{Autonomous robots}, vol.~15, no.~2, pp. 111--127, 2003.

\bibitem{Thrun:C01}
S.~{Thrun}, ``Learning occupancy grids with forward models,'' in \emph{Proc.
  IEEE/RSJ Int. Conf. on Intelligent Robots and Syst. Expanding the Societal
  Role of Robotics in the the Next Millennium (Cat. No.01CH37180)}, vol.~3,
  2001, pp. 1676--1681 vol.3.

\bibitem{robbiano2019bayesian}
C.~Robbiano \emph{et~al.}, ``Bayesian learning of occupancy grids,''
  \emph{arXiv preprint arXiv:1911.07915}, 2019.

\bibitem{guidi2016joint}
F.~Guidi \emph{et~al.}, ``Joint energy detection and massive array design for
  localization and mapping,'' \emph{IEEE Trans. Wireless Commun.}, vol.~16,
  no.~3, pp. 1359--1371, 2016.

\bibitem{MarGioChi:J11}
A.~{Mariani}, A.~{Giorgetti}, and M.~{Chiani}, ``Effects of noise power
  estimation on energy detection for cognitive radio applications,'' \emph{IEEE
  Trans. Commun.}, vol.~59, no.~12, pp. 3410--3420, Dec. 2011.

\bibitem{mafi2011information}
N.~Mafi, F.~Abtahi, and I.~Fasel, ``Information theoretic reward shaping for
  curiosity driven learning in pomdps,'' in \emph{Proc. IEEE Int. Conf.
  Develop. Learning}, vol.~2, 2011, pp. 1--7.

\end{thebibliography}

\end{document}